Article

# Towards Context-Rich Automated Biodiversity Assessments: Deriving AI-Powered Insights from Camera Trap Data

**Paul Fergus [1,*], Carl Chalmers [1], Naomi Matthews [2], Stuart Nixon [2], André Burger [3], Oliver Hartley [4], Chris Sutherland [4], Xavier Lambin [5], Steven Longmore [6] and Serge Wich [7]**

[1] School of Computer Science and Mathematics, Liverpool John Moores University, James Parsons Building, Byrom Street, Liverpool, L3 3AF, UK; c.chalmers@ljmu.ac.uk
[2] Chester Zoo, Upton-by-Chester, Chester, CH2 IEU, UK; n.matthews@chesterzoo.org; s.nixon@chesterzoo.org
[3] Welgevonden Game Reserve, P.O. Box 433, Vaalwater 0530, South Africa; andre@welgevonden.org
[4] School of Mathematics and Statistics, University of St Andrews, Mathematical Institute, North Haugh, St Andrews, KY16 9SS, Scotland, UK; oh57@st-andrews.ac.uk; css6@st-andrews.ac.uk
[5] School of Biological Sciences, University of Aberdeen, Tillydrone Avenue, Aberdeen, AB24 2TZ, Scotland, UK; x.lambin@abdn.ac.uk
[6] Astrophysics Research Institute, Liverpool John Moores University, IC2, Liverpool Science Park, 146 Brownlow Hill, Liverpool, L3 5RF, UK; s.n.longmore@ljmu.ac.uk
[7] School of Biological and Environmental Sciences, Liverpool John Moores University, James Parsons Building, Byrom Street, Liverpool, L3 3AF, UK; s.a.wich@ljmu.ac.uk
* Correspondence: p.fergus@ljmu.ac.uk

**Abstract:** Camera traps offer enormous new opportunities in ecological studies, but current automated image analysis methods often lack the contextual richness needed to support impactful conservation outcomes. Here we present an integrated approach that combines deep learning-based vision and language models to improve ecological reporting using data from camera traps. We introduce a two-stage system: YOLOv10-X to localise and classify species (mammals and birds) within images, and a Phi-3.5-vision-instruct model to read YOLOv10-X binding box labels to identify species, overcoming its limitation with hard to classify objects in images. Additionally, Phi-3.5 detects broader variables, such as vegetation type, and time of day, providing rich ecological and environmental context to YOLO's species detection output. When combined, this output is processed by the model's natural language system to answer complex queries, and retrieval-augmented generation (RAG) is employed to enrich responses with external information, like species weight and IUCN status (information that cannot be obtained through direct visual analysis). This information is used to automatically generate structured reports, providing biodiversity stakeholders with deeper insights into, for example, species abundance, distribution, animal behaviour, and habitat selection. Our approach delivers contextually rich narratives that aid in wildlife management decisions. By providing contextually rich insights, our approach not only reduces manual effort but also supports timely decision-making in conservation, potentially shifting efforts from reactive to proactive management.

## 1. Introduction

Camera traps have proven useful in wildlife conservation efforts, offering insights into animal population dynamics and habitat use across large geographical areas without the need for direct human observation [1], [2]. Through image and video data collection these motion-triggered devices provide novel, and at times inaccessible data by other means for biodiversity assessment and the evaluation of conservation strategies [3]. However, the sheer volume of data being generated by camera trap projects presents unique challenges [4]; environmental factors, such as moving vegetation or changing light, often lead to false positives [5], adding noise that complicates species identification and demands additional resources to manage [6] [7].

These challenges make processing and analysing the datasets particularly complex [8]. Traditionally, the initial stages of camera trap image analysis involve manual annotation, often conducted by experts or through citizen science initiatives [9], [10]. This annotation process, which converts images into a structured format such as CSV for further analysis, is time-consuming [11], costly, and prone to human error [12]. Subsequent analysis typically requires additional software tools, such as R [13], and a deeper understanding of statistical and special methodologies to derive meaningful insights.

In relation to manual annotation limitations, automated detection models, such as MegaDetector [14], have been developed to assist in identifying general categories like animals, humans, and vehicles in images [15]. By reducing the need for manual review, MegaDetector significantly helps to improve workflow efficiency, making it an essential tool in ecological studies, [16], [17]. However, while effective at broad categorisation, MegaDetector lacks species-specific identification, limiting its utility in detailed ecological assessments [18], [19].

To address this issue, advances in object detection models [20], such as YOLO (You Only Look Once), have introduced improved capabilities for species-specific identification [21], [22], making them increasingly applicable to camera trap data [8], [23]. Constrained by the architecture of the model and its tailored training set, it is however unable to detect anything outside of what it has been trained on. This is a significant limitation in situations where context-rich information is needed [18], such as animal behaviours (e.g., sitting, standing, feeding) and environmental context (e.g., habitat damage [24] or the presence of invasive plants) [23].

Recognising the critical role of contextual understanding in enhancing detection accuracy, recent research has increasingly focused on multimodal large language models (MLLMs) [25], such as ContextDET, which integrate cues from human-AI interactions to improve object detection in complex scenes [26]. ContextDET, a vision language model (VLM) [27] utilises a generate-then-detect framework, combining visual encoders, pre-trained language models, and visual decoders to locate and identify objects within diverse contextual environments, effectively responding to open-ended queries [28]. Building on these advancements, models like VCoder introduced versatile vision encoders, designed specifically to enhance object perception tasks, such as counting or identifying entities within cluttered scenes, where traditional MLLMs may struggle [29]. Meanwhile, VisionLLM offers a flexible, open-ended task approach by aligning language instructions with vision tasks, which enables a range of vision-centric functions like object detection, image captioning, and visual reasoning [30]. Although these frameworks have shown success in urban applications, their adaptation for conservation remains limited, presenting a valuable opportunity to leverage their contextual capabilities for complex wildlife monitoring environments [31].

In response to this need, we propose an integrated approach that combines deep learning-based vision and language models to enhance camera trap image understanding [32]. Specifically, our method merges the object detection capabilities of YOLOv10-X [33] with the vision-language understanding of Microsoft's Phi-3.5-vision-instruct transformer model [34], [35]. In addition, our system integrates a Retrieval-Augmented Generation (RAG) framework [36], allowing it to draw on external sources, such as the IUCN Red List [37], for answering complex queries about camera trap images. While existing tools, such as the R package "traitbase" [38] and platforms like the Open Traits Network [39], offer the ability to attach additional trait characteristics (e.g., average weight, IUCN status) to species lists, they often require users to have programming expertise and domain knowledge. In contrast, our system automates this process allowing users to obtain insights through natural language queries, such as, "*What species is in this image, how much does an average adult weigh, and what is its IUCN status?*". By integrating these capabilities, our approach aims to simplify workflows and provide faster access to actionable data, making advanced tools accessible to a broader audience, including non-specialist users.

Building on prior detection and contextual models [21], [22], our approach generates structured, context-aware outputs that consolidate species-specific and environmental insights from camera trap data. While these outputs provide foundational information, such as species identification, estimated counts, and behavioural or habitat context, further analysis is still necessary to interpret these insights for specific stakeholder needs. This could include translating the data into biodiversity metrics, such as biocredit trading [40], or monitoring environmental changes, like habitat destruction or poaching activities (for example through weapon detection) [30], [41]. By streamlining the annotation process and providing immediate insights, our approach has the potential to enhance conservation efforts and resource allocation [42], [43]. However, we acknowledge that the full impact of this system will depend on integrating its outputs into broader conservation workflows and decision-making processes.

The remainder of this paper expands on this approach. Section 2 details our Methodology, presenting the innovative solution we developed. Section 3 presents the results, followed by a discussion in Section 4, with final conclusions and future directions in Section 5.

## 2. Methodology

In this section, we present the methodology to enhance the analysis of camera trap images. Our system integrates object detection, vision-language modelling, and retrieval-augmented generation (RAG) to deliver detailed, context-rich reports of wildlife in natural habitats. By combining these complementary components, our approach addresses the limitations of traditional methods, offering a more comprehensive and automated tool for environmental monitoring. The methodology consists of four broadly defined stages: object detection for animal identification, vision-language modelling for visual understanding and question answering, and RAG for contextual enrichment using external knowledge.

Below, we provide a system overview before discussing each stage in detail and describing the specific configurations and processes used to achieve our objectives.

### *2.1. System Overview*

Figure 1 shows the workflow designed to automate the transformation of camera trap images into actionable insights. The process begins with the ingestion of camera trap images (Step 1), which are analysed using YOLOv10-X to detect and spatially localise animal species (Step 2). Contextual analysis is conducted via the Phi-3.5 model, which reads the binding box labels generated in Step 2 and extracts additional domain-specific information from other identified objects known to the model. The outputs from these models are then integrated in Step 4, producing a comprehensive dataset optimised for downstream analysis and interpretation.

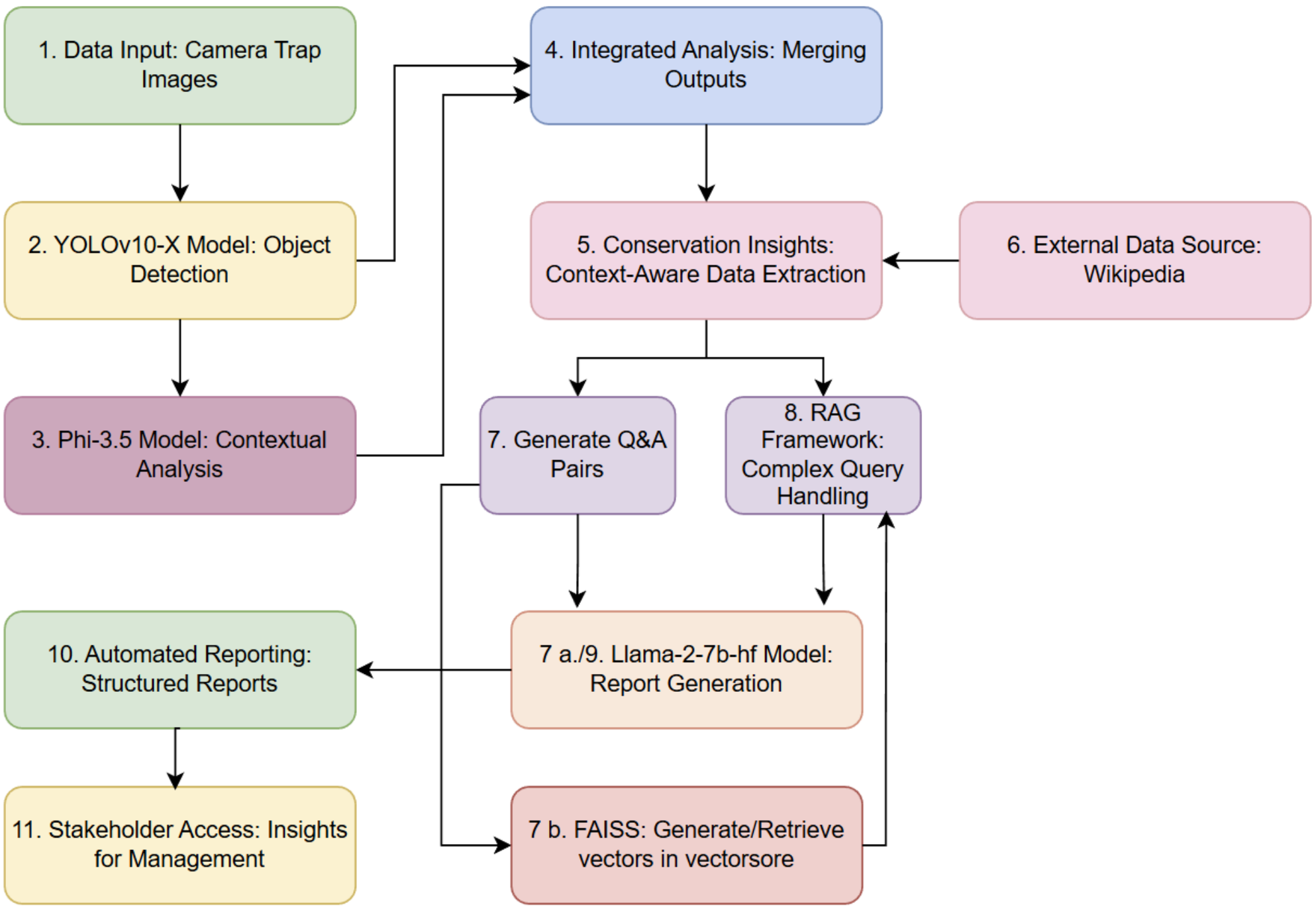

**Figure 1.** Flow chart illustrating an overview of the workflow for the multi-modal YOLOv10-x and Phi3.5 systems integration for context-rich camera trap data processing.

Conservation-specific insights are extracted with contextual awareness, further enriched by supplementary information retrieved from external sources, such as the IUCN Red List (Step 5 and 6). This system generates structured question-and-answer pairs, facilitating systematic information retrieval and report generation (Step 7, 7a, 7b and 8). Complex queries are addressed through a RAG framework, which leverages this contextual information to deliver precise and relevant responses. Insights are compiled into structured reports using the Llama-2-7b-hf model [44], which formats the outputs for direct accessibility by stakeholders (Step 10 and 11). Additionally, a FAISS-based vector store [45] underpins efficient query management, enabling rapid retrieval and real-time access to critical insights.

### *2.2. Data Collection*

By collaborating with global conservation organisations, Conservation AI [21] has compiled diverse camera trap datasets that represent a wide range of habitats. This diversity in species and environments ensures that the AI models developed are robust and adaptable across ecosystems, significantly enhancing their utility for global conservation efforts.

For this study, we utilise our Sub-Saharan Africa dataset, which contains 57,120 tagged objects across 41,111 RGB images representing 31 distinct classes (29 animal species, including a person and car class), as shown in Figure 2. These camera trap images were sourced from across southern and central-African regions. High-quality image tagging, performed by specialists and managed through our in-house quality control process, ensures precise bounding box annotations. This consistency is essential for optimising model performance, given the complexity and scale of the dataset.

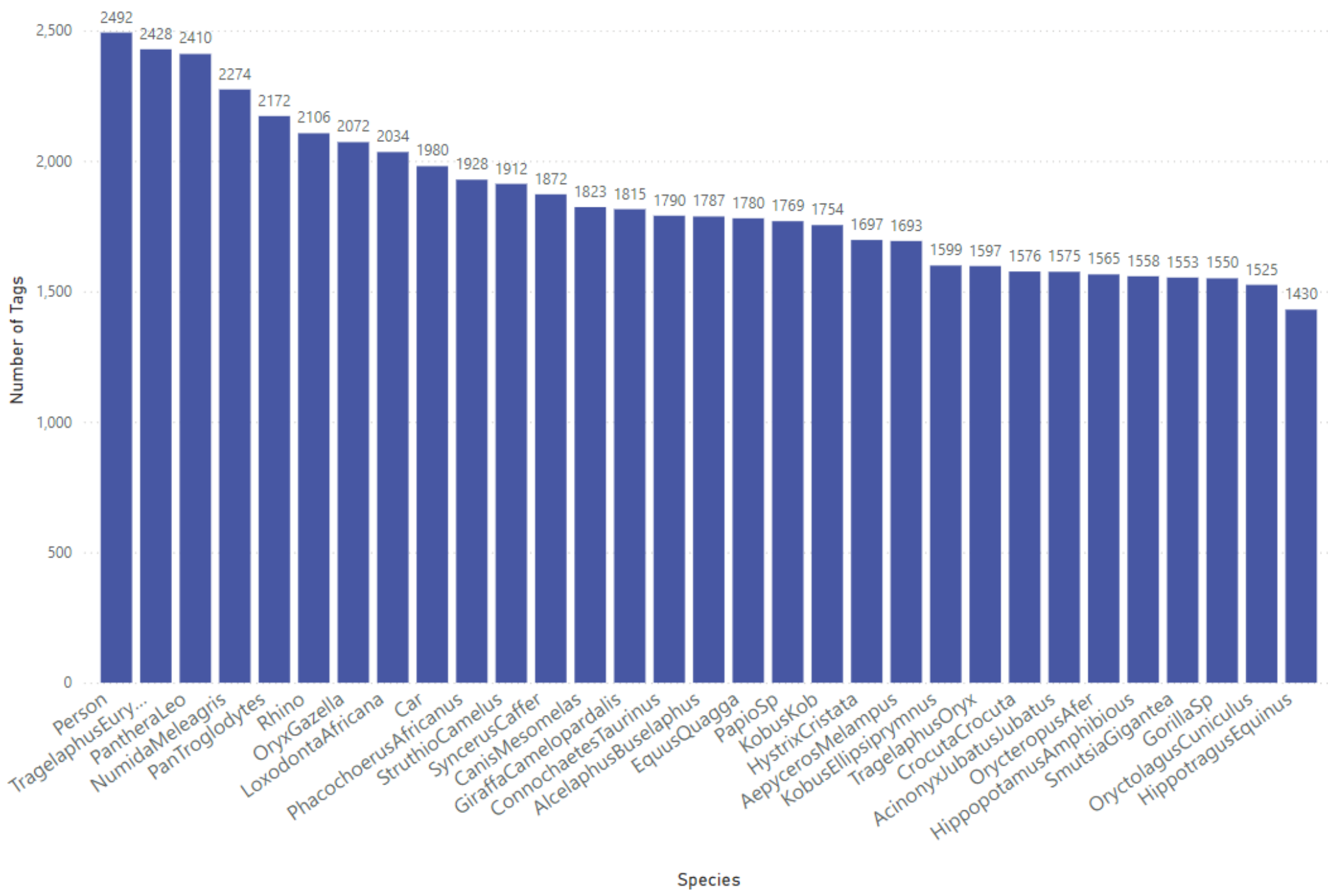

**Figure 2.** Class distribution for the sub-Saharan Africa dataset used to train the YOLOv10-X model to localise and detect mammals, birds, people and cars.

The dataset is divided into training, validation, and test sets with an 80:10:10 split. The training set facilitates model learning, the validation set is used for hyperparameter tuning, and the test set evaluates performance using mean average precision (mAP) and intersection over union (IoU) metrics (See Section 2.8) [46]. This split ensures the model generalises well to unseen data and avoids overfitting (overfitting is when the model becomes too tailored to its training data, sometimes fitting it exactly, which leads to a model that struggles to make accurate predictions or inferences on data outside the training set). Additionally, 602 independent camera trap images were collected to evaluate the YOLOv10-X object detection model and the Phi-3.5-vision-instruct model's ability to identify additional objects outside the capabilities of our YOLOv10-X model.

### *2.3. Object Detection Model*

YOLOv10-X [33] is well-suited for handling challenging real-world imagery, such as low-quality or obscured camera trap images. It outperforms our previous custom trained Faster Region Convolutional Neural Network (FasterRCNN) model [47], [48] in both detection accuracy and speed [22]. Using the Microsoft COCO dataset [49], YOLOv10-X was trained using 8 NVIDIA RTX 3090 GPUs over a 10-day period, with additional transfer learning on our Sub-Saharan Africa wildlife dataset (took approximately 3 hours using 8 RTX A6000 GPUs using a batch size of 256).

The YOLOv10 architecture (Figure 3) incorporates an enhanced version of CSPNet (Cross Stage Partial Network) [50] to improve gradient flow and reduce computational redundancy, making it highly efficient for large-scale datasets. The CSPNet backbone extracts key features from input images, crucial for handling the variability seen in wildlife camera trap imagery. To ensure robustness across conditions, the model's neck uses Path Aggregation Network (PAN) layers for effective multiscale feature fusion, enabling the detection of animals of various sizes, from small birds to large mammals [51]. During training, the model uses a one-to-many head to generate multiple predictions per object, enhancing learning accuracy. In the inference phase, it shifts to a one-to-one head, eliminating the need for Non-Maximum Suppression (NMS) and reducing latency [52].

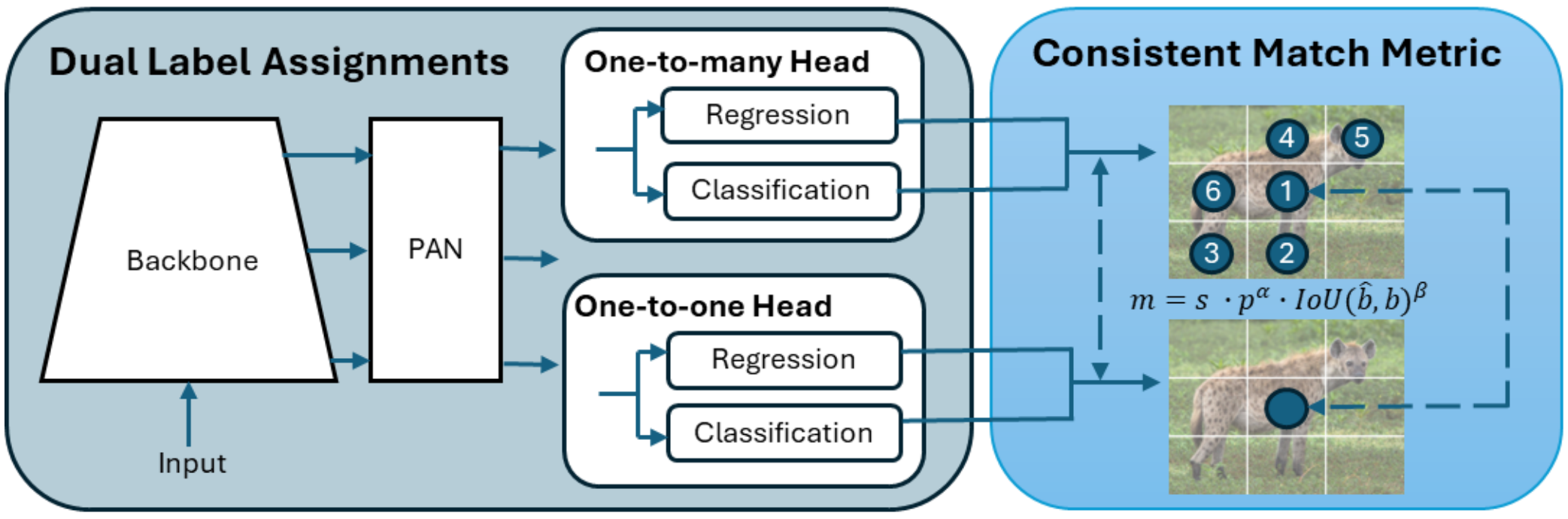

**Figure 3.** Overview of the YOLOv10 architecture.

YOLOv10X incorporates lightweight classification heads, spatial-channel decoupled downsampling, and rank-guided block design, which reduce computational overhead without compromising accuracy. Large-kernel convolutions [53] and partial self-attention modules [54] further enhance its ability to process complex scenes, without increasing computational cost. These optimisations ensure that YOLOv10-X offers a powerful, efficient solution for wildlife monitoring where speed and accuracy are crucial. The model is deployed on an NVIDIA Triton Inference Server, providing serverless API endpoints for easy integration without requiring extensive infrastructure [55].

As input images are processed, YOLOv10-X outputs bounding boxes and labels for detected animals, isolating them for further analysis (see Figure 4). These labelled images (complete image with binding box) are then passed to the Phi-3.5-vision-instruct model to extract additional object details and contextual information. Since the Phi-3.5 model struggles to detect animals in low-quality images, it reads the species names from the bounding box labels. Fine-tuning the Phi-3.5 model to improve performance in such cases would require significant hardware resources, so we adopted this hybrid approach. By using YOLOv10-X for detection and classification and the Phi-3.5 model for context analysis, we efficiently leverage the strengths of both models. The Phi-3.5 model also detects environmental details, such as trees, weather, and embedded metadata text (e.g., 25/05/2022 05:29:28 WED) (Figure 4). This multi-step process ensures the system provides comprehensive analysis while remaining computationally feasible.

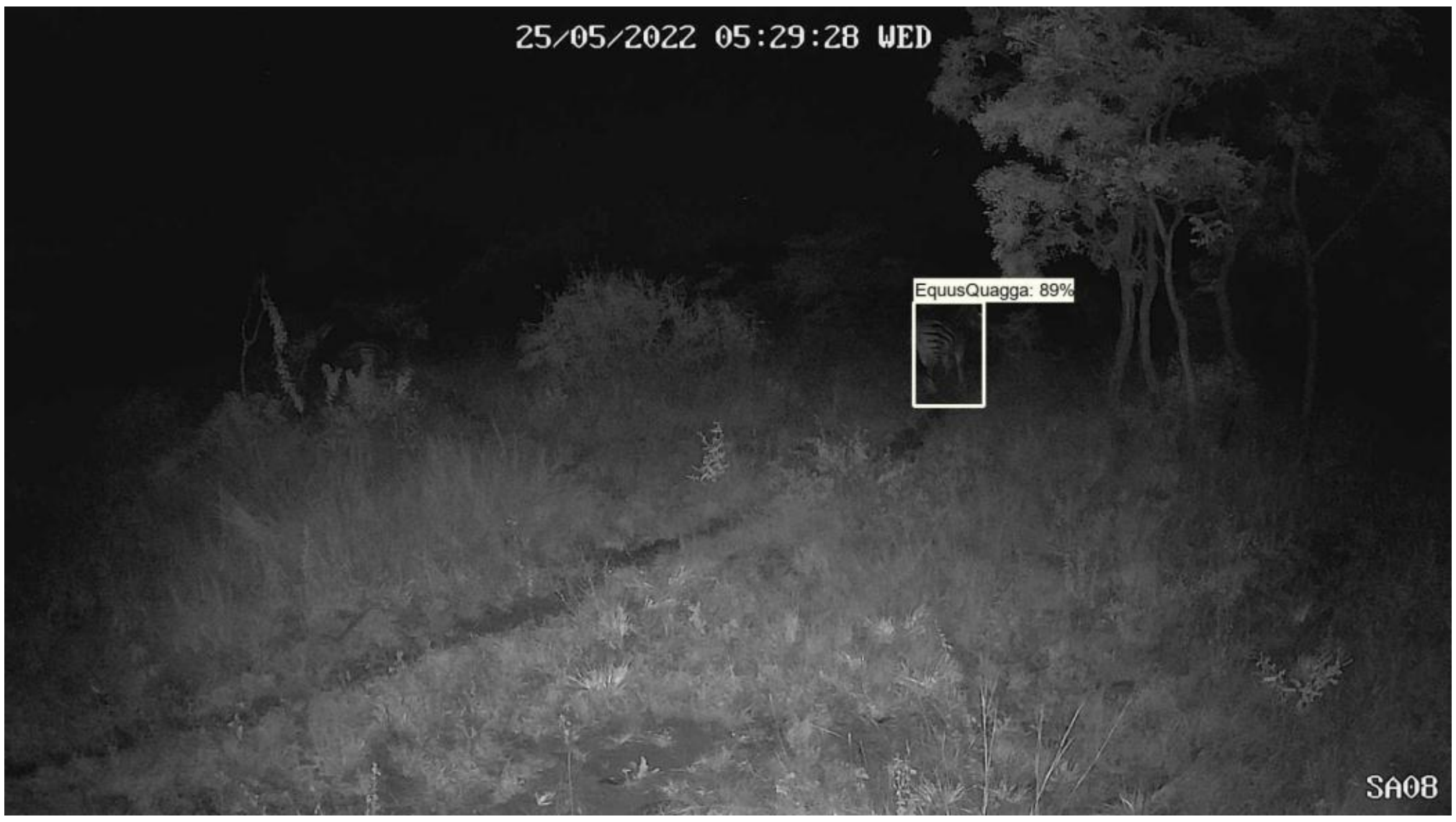

**Figure 4.** Image from Limpopo Province in South Africa show the detection of a Zebra at night using a camera trap.

*2.4. Vision Language Model*

The Phi-3.5-vision-instrut model is used to analyse the YOLOv10-X classified images [56]. As a state-of-the-art multimodal system, it processes both text and image data, making it ideal for tasks requiring a deep understanding of visual content and contextual information. The model is equipped with an image processor that handles up to 16 crops of the input image, allowing it to focus on different regions for more detailed analysis. Efficient resource allocation ensures optimal use of hardware, such as GPUs, to handle the large volume of images encountered in wildlife monitoring projects.

The Phi-3.5-vision-instruct model consists of 4.2 billion parameters, enabling it to efficiently manage large-scale data and complex tasks. With a context length of up to 128K tokens, the model can handle extensive sequences of text and visual data, allowing for the generation of rich, detailed descriptions from both image and textual content. Trained on 500 billion tokens using 256 A100-80G GPUs over six days, the model's training regime ensures high accuracy and adaptability across diverse contexts. The model's backbone extracts features from both text and images, leveraging advanced neural network architectures to capture complex relationships within the data. Like YOLOv10, the Phi-3.5 model integrates large-kernel convolutions [57] and partial self-attention modules [58], enhancing its ability to focus on relevant parts of the input data [53], [54]. This shared architectural optimisation across models ensures efficient processing without compromising on accuracy. Similar to YOLOv10, the Phi-3.5-vision-instruct model is deployed on a NVIDIA Triton Inference Server, ensuring efficient integration into the overall system.

The Phi-3.5 model excels in tasks requiring a deep understanding of both textual and visual data. However, when applied to camera trap images, such as those shown in Figure 4, the model struggles to reliably detect and identify specific species. To address this limitation, we rely on the fine-tuned YOLOv10 model to first detect and classify animals (because it is much easier and cheaper to train a YOLO model than a Phi model), placing labelled bounding boxes around each detected animal. The Phi-3.5-vision-instruct model then reads these bounding box labels to identify the species, and supplements this with additional contextual information it can detect in the image. This hybrid approach leverages the strengths of both models to ensure accurate and context-rich analysis, even in challenging conditions.

Using Figure 4 as an example, the Phi-3.5-vision-instruct model identifies the species as *Equus quagga* (zebra), the camera ID as *SA08*, and the time and date as *25/05/2022, 05:29:28 WED*. It also infers that the image was taken in the dark in a wooded environment based on the presence of trees and grass. This contextual information, combined with object detection results, provides a more complete understanding of the scene. In Figure 5, the model detects four *Connochaetes taurinus* (Blue wildebeest) and two *Equus quagga* (Pains zebra), offering useful data for species abundance estimation or population dynamics.

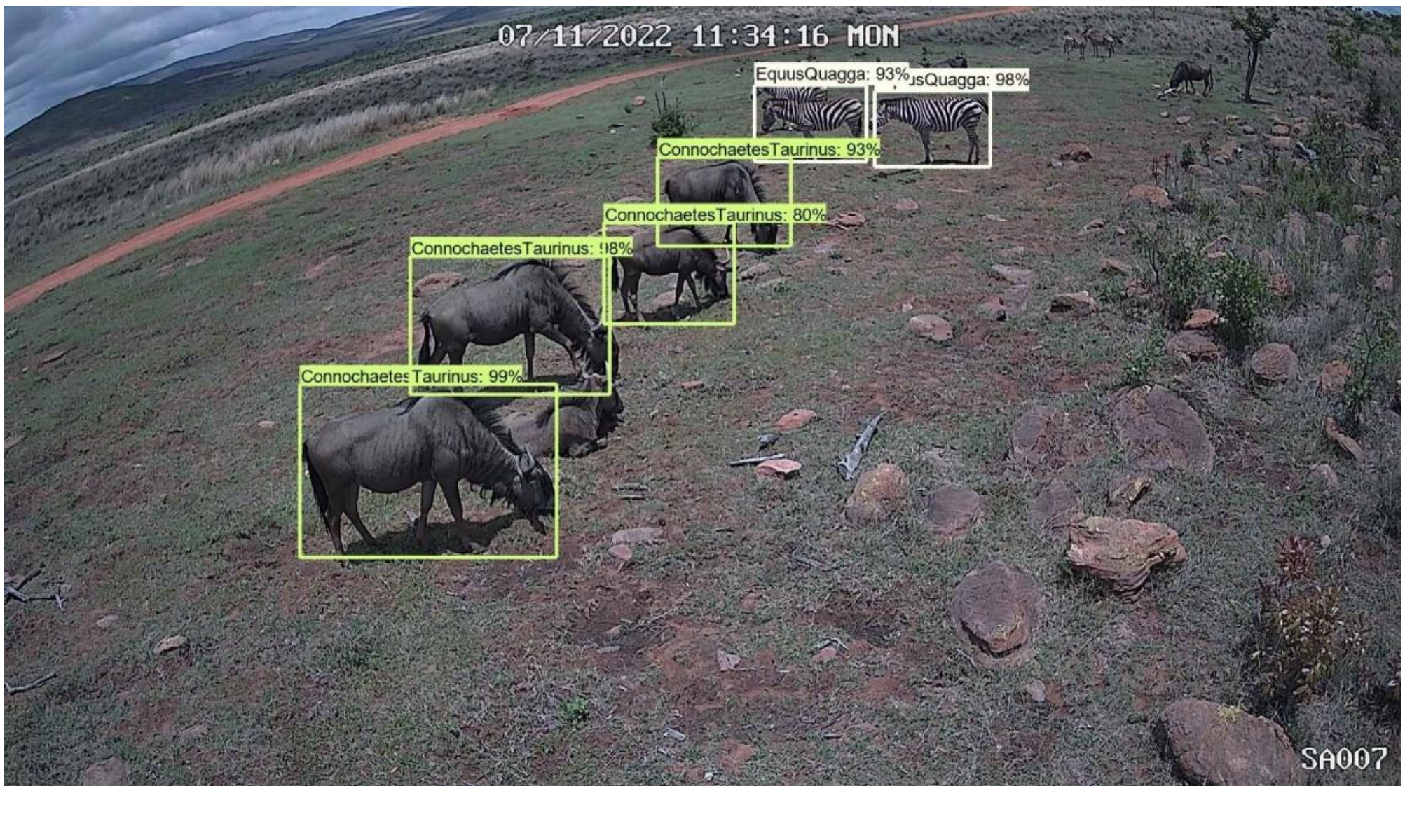

**Figure 5.** Image from Limpopo Province in South Africa showing the detection of a multiple Blue Wildebeest and Zebras using a camera trap.

For information not directly available from the image - such as species weight, IUCN status, or other biological facts - the system integrates RAG to retrieve knowledge from external sources. This ensures the system delivers not only image-based insights but also enriched external data, enhancing its overall utility for conservationists.

### *2.5. Retrieval-Augmented Generation (RAG)*

The RAG component is implemented using LangChain [59] and sources external information from Wikipedia based on image-extracted data (we use Wikipedia because large scale scrapping of the IUCN Red List is forbidden resulting in revoked tokens). For example, when queried with "*What animal is in this image, how much does it weigh, and what is its IUCN status?*", the system generates relevant keywords from the Phi-3.5-vision-instruct model, which are then used to search for Wikipedia documents. The retriever gathers up to 20 Wikipedia pages, which are split into smaller passages for precise extraction. Facebook AI Similarity Search (FAISS) hen ranks these passages based on relevance [45]. By controlling the number of retrieved passages and the chunk size, the system ensures that the most relevant information is selected. Combining this external knowledge with image analysis provides the basis for vision question answering.

### *2.6. Visual Question Answering*

During the visual question answering stage, the system generates comprehensive answers to questions by integrating visual analysis with the retrieved external knowledge. The information from the retrieved passages is combined with the visual analysis, resulting in detailed, contextually enriched responses. The system returns a tuple (an array containing the final answer, the keywords used for retrieval, and the selected Wikipedia passages), ensuring transparency in how the answer was formulated. This transparency is crucial for researchers, developing a trust in the system's outputs.

### *2.7. Automatic Reporting*

The final stage in the methodology includes the integration of an automated reporting system. Once the images are processed by the object detection and vision-language models, the system uses the Alpaca format to generate structured reports [60]. The Llama-2-7b-hf model automatically formulates relevant questions based on the extracted data, and the system provides answers using outputs from the Phi-3.5 model – this process converts unstructured data (response from the Phi-3.5 model) into a structured set of questions and answers in a Alpaca JSON format [61] for better post-processing and report generation. The reports offer comprehensive summaries of the findings, that significantly reduces the need for manual report generation, ensuring that detailed, consistent information is readily available as and when it is needed.

### *2.8. Evaluation Metrics*

To assess the performance of the YOLOv10-X and Phi-3.5-vision-instruct models, we employ several key metrics, including precision, recall, F1-score, mAP, IoU, and BERTScores for evaluating generated answers against ground truth answers. *Precision* measures the proportion of true positives (correctly identified animals) among all positive predictions, reflecting the model's classification accuracy. Recall evaluates the model's ability to identify all relevant animals within images, calculating the proportion of true positives out of the total actual number of animals present. The F1-score balances precision and recall, providing a more comprehensive evaluation, particularly in scenarios where minimising false positives is crucial.

mAP is a key metric in object detection, measuring the average *precision* across all classes at various IoU thresholds. IoU quantifies the overlap between the predicted and actual bounding boxes, with a high IoU value indicating more accurate localisation. mAP provides a comprehensive evaluation of the model's ability to detect and label animals accurately.

For evaluating the system's answering capabilities, we use BERTScore, which measures precision, recall, and F1 to assess the quality of generated and ground truth answers [62]. This metric determines how closely the generated answers align with expected responses, particularly when visual data is enriched with external knowledge.

Each of these metrics contributes to a detailed understanding of the models' strengths and weaknesses, ensuring a thorough evaluation across species detection, classification, and contextual information retrieval tasks.

## 3. Results

The results in this section are structured around the system's multi-stage approach, which integrates object detection, VLM modelling, and RAG to deliver detailed, contextually rich descriptions of wildlife. Each component is evaluated based on its accuracy, robustness, and contribution to the overall system's effectiveness.

### *3.1. Training Results for the Sub-Saharan Model*

The YOLOv10-X model was trained to detect and classify 29 species in Sub-Saharan Africa, along with vehicles and human subjects. The dataset includes a diverse range of fauna, such as *Acinonyx jubatus* (cheetah), *Panthera leo* (lion), and *Loxodonta africana* (African elephant), presenting challenges due to the variation in morphology, size, and behaviour among species. The Precision-Recall (PR) curve (Figure 6) shows a mAP of 0.976 at a 0.5 IoU threshold, reflecting high detection accuracy across all classes. Precision-Recall curves are essential for evaluating object detection tasks, as they illustrate the trade-off between detecting all relevant objects (recall) and avoiding false positives (precision). Additionally, F1-Confidence curves and confusion matrices provide valuable insight into the model's performance across various confidence thresholds and help identify misclassifications across species.

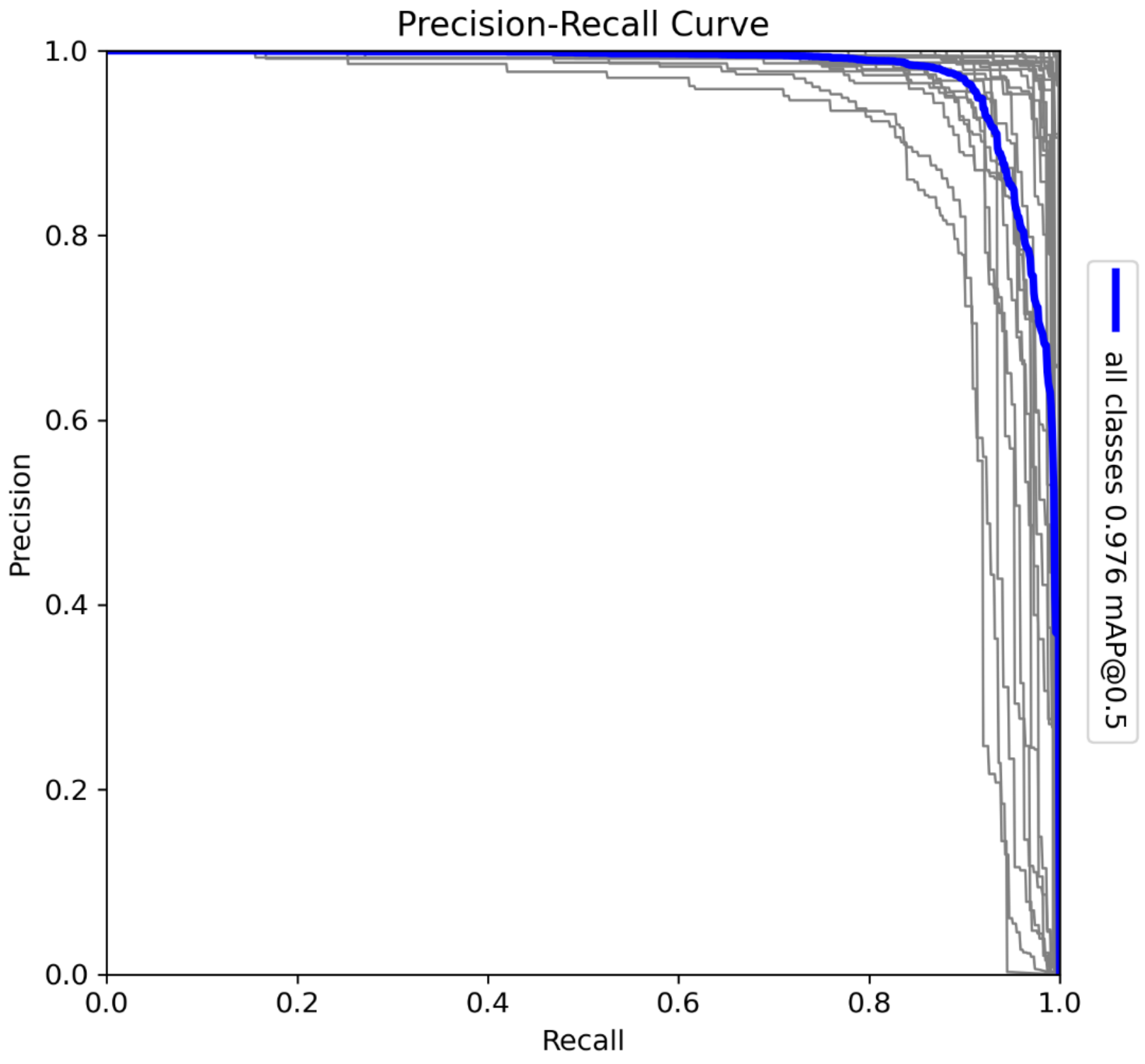

**Figure 5.** Precision-Recall (PR) curve for the YOLOv10-X model trained on 29 Sub-Saharan African species, vehicles, and human subjects.

The Precision-Confidence curve (Figure 7) provides insights into the reliability of the YOLOv10-X model's predictions across all classes. The curve demonstrates that the model achieves high precision even at low confidence thresholds, with predictions remaining accurate at moderate confidence levels. At the maximum level of 1.0, the model achieves perfect precision. Variations in individual class curves, particularly for more visually similar species, indicate classification challenges, but the strong correlation between precision and confidence confirms the model's reliability in making accurate detections across different confidence levels.

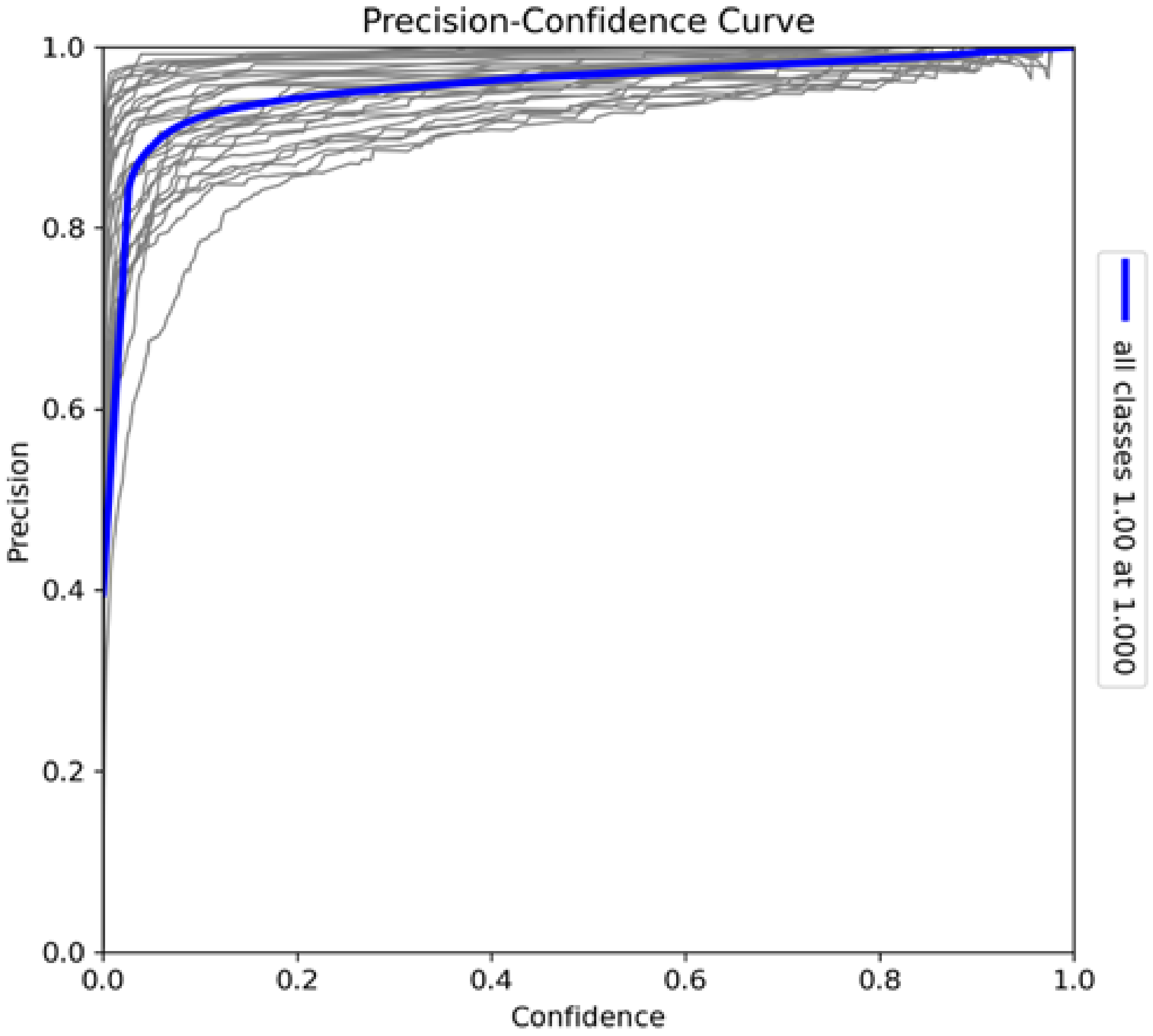

**Figure 7.** Precision-Confidence curve for the model trained on Sub-Saharan African species, vehicles, and human subjects.

The Recall-Confidence curve (Figure 8) illustrates the trade-off between recall and confidence thresholds. At lower thresholds, the model achieves near-perfect recall (0.99), indicating that it captures almost all true positives when not constrained by confidence. However, as the confidence threshold increases, recall declines, with a sharp drop near the highest confidence levels, where precision is prioritised. Variations in individual species curves suggest that certain species may benefit from more relaxed thresholds to improve recall (particularly species such as *Bunolagus monticularis* (Bushman Rabbit) which are more difficult to detect than larger animals). Despite this trade-off, the model demonstrates strong overall recall performance, ensuring comprehensive detection coverage at lower confidence thresholds.

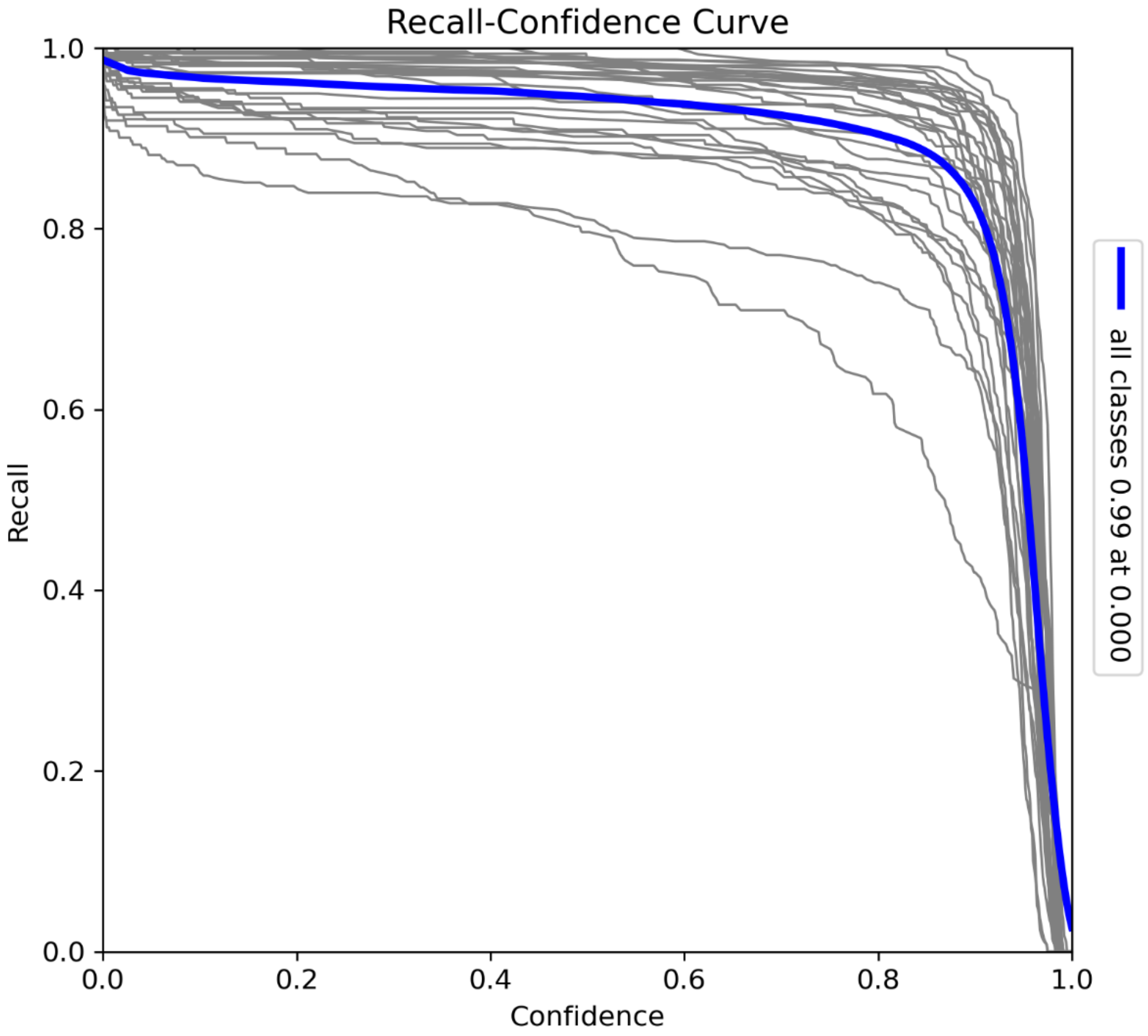

**Figure 8.** Recall-Confidence curve for the model trained on Sub-Saharan African species, vehicles, and human subjects.

The F1-Confidence curve (Figure 9) provides a comprehensive evaluation of the model's balance between precision and recall across various confidence thresholds. The F1-score peaks at 0.96 when the threshold is 0.422, indicating optimal performance at this level. While the score remains high over a broad range of confidence levels, a sharp decline occurs near the maximum confidence threshold due to missed instances as recall decreases. Nevertheless, the model consistently achieves a strong F1-score, highlighting its effectiveness in balancing precision and recall for accurate detection tasks.

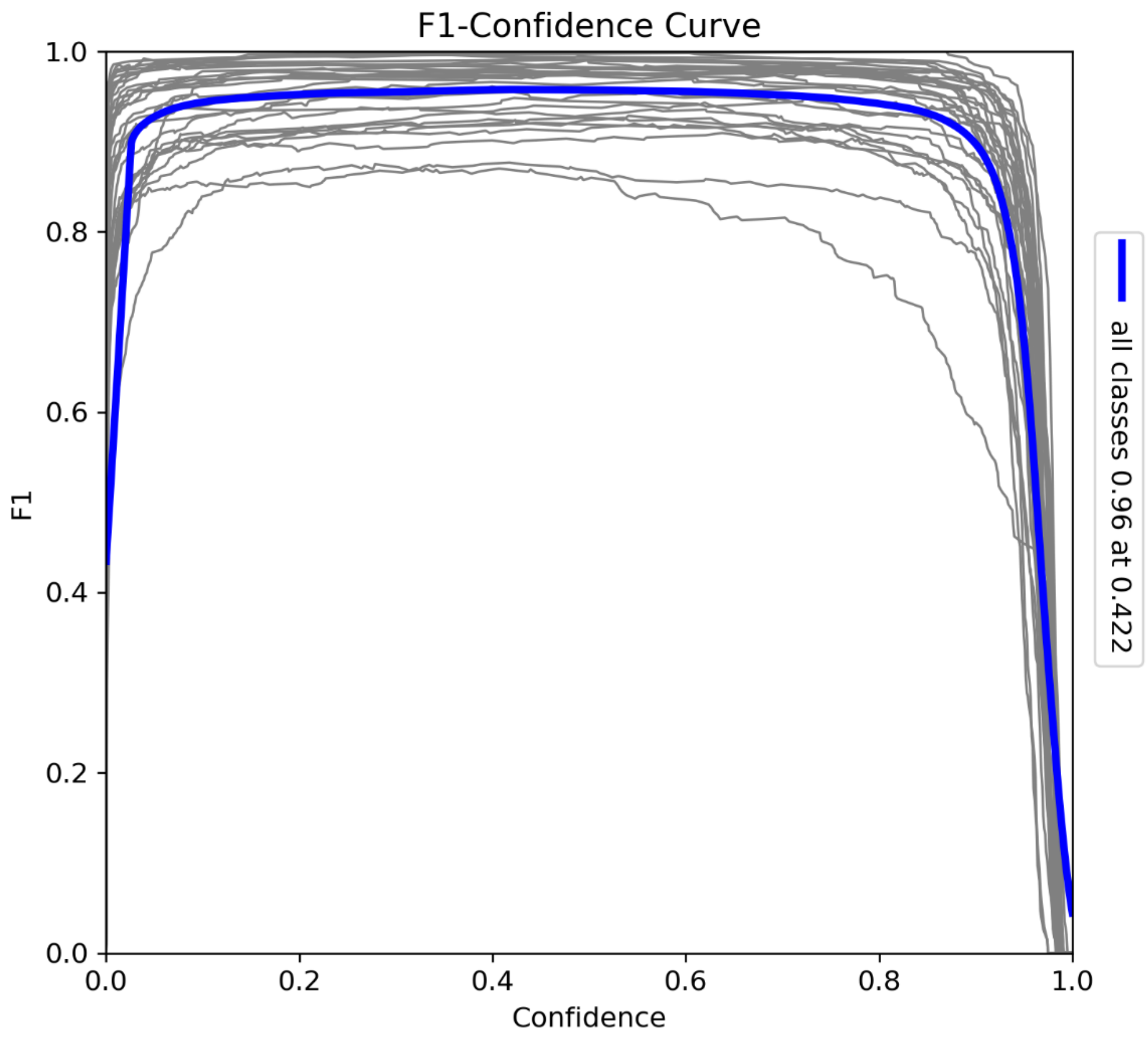

**Figure 9.** F1-Confidence curve for the model trained on Sub-Saharan African species, vehicles, and human subjects.

The confusion matrix (Figure 10) provides a detailed analysis of the model’s classification performance across the Sub-Saharan African species dataset. A strong diagonal indicates correct predictions, with high values showing accurate identification of most species. The highest frequencies are observed for commonly occurring classes like *Acinonyx jubatus* (Cheetah) and *Loxodonta Africana* (African elephant), reflecting the model’s effectiveness with these species. Off-diagonal cells represent minimal misclassifications, likely due to visual similarities between species. Overall, the matrix confirms the model's strong classification accuracy across the majority of classes, with relatively low rates of misclassification.

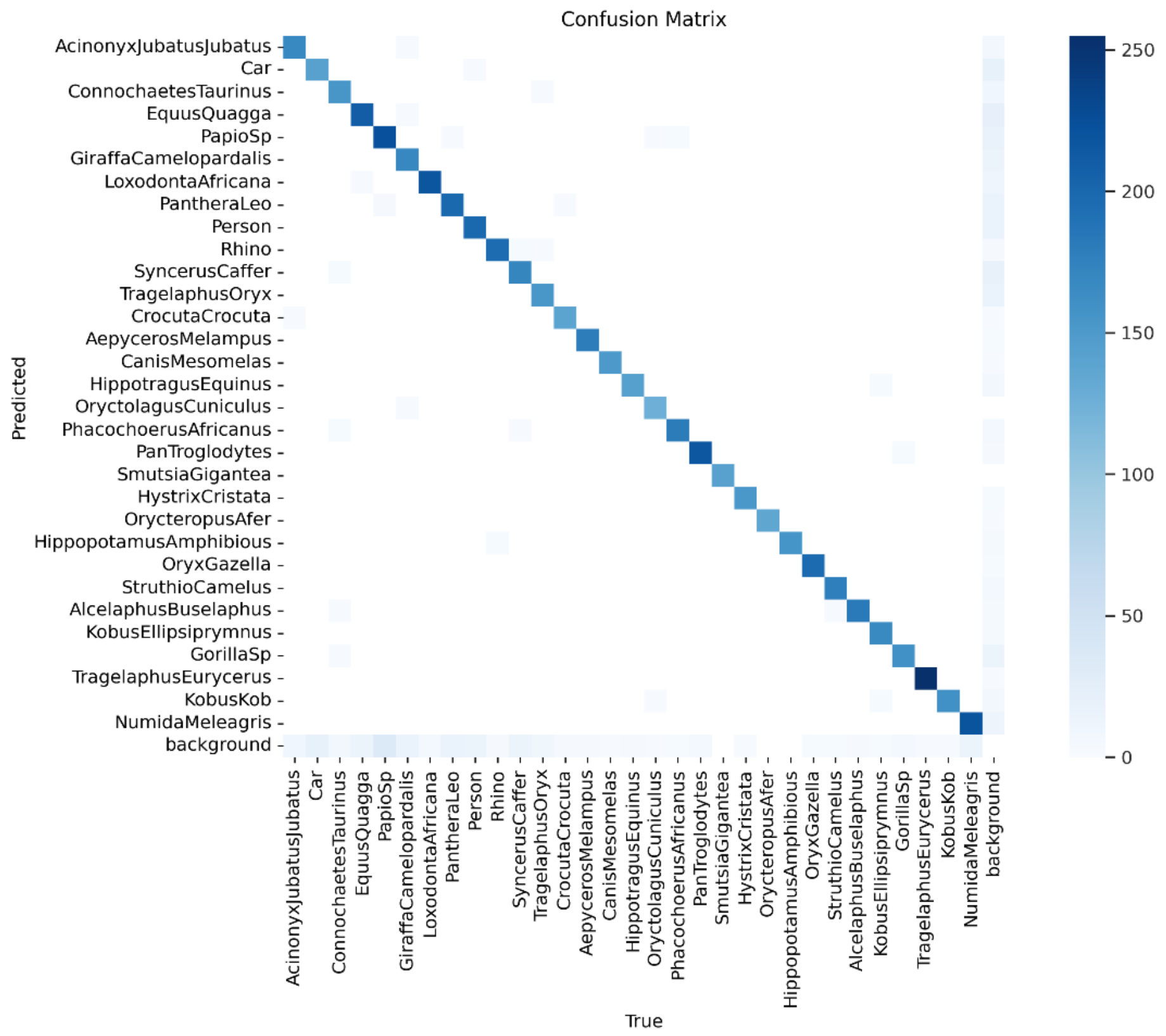

**Figure 10.** The confusion matrix provides a detailed analysis of the model's classification performance across all Sub-Saharan African species, vehicles, and human subjects.

### 4.2. Results for Vision-Language Model Without YOLOv10-X Object Detection Support

This section presents the performance results for the Phi 3.5-vision-instruct model when applied directly to the 602 independent camera trap images without object detection support. In this scenario, the model relies solely on its vision-based capabilities to detect and classify animals. As shown in Table 1, the model demonstrates high precision across most classes, with some achieving perfect precision (1.0000). However, the recall values are significantly lower for several species, indicating difficulties in identifying all instances of the species. Despite overall accuracy exceeding 90% in most cases, the recall and F1-score metrics highlight areas where the model struggles with complete identification.

Class-wise, the model's performance varies significantly. For example, *Canis mesomelas* (Black-backed jackal) achieves an accuracy of 98.33%, but its low recall (0.2000) results in an F1-score of 0.3333, highlighting difficulties in consistently detecting this species. In contrast, *Syncerus caffer* (African buffalo) (recall 0.6364) and *Struthio camelus* (Common ostrich) (recall 0.6000) show stronger performance, with F1-scores of 0.7778 and 0.7500, respectively. Similarly, *Gorilla sp* (Gorilla) performs well, with a recall of 0.6818 and an F1-score of 0.8108, demonstrating the model's effectiveness in identifying more visually distinct species. It is important to note that some animals, such as gorillas, are grouped at the genus level (*Gorilla sp.*) rather than by subspecies. This is because object detection models often struggle to differentiate between closely related subspecies due to their high visual similarity. For example, distinguishing between *Gorilla beringei* (Easter gorilla) and *Gorilla gorilla* (Western gorilla) would require finer-grained visual features than those typically captured in camera trap data. As a result, grouping by genus ensures more reliable detection and avoids introducing additional errors into the analysis.

The model struggles with certain species, particularly *Rhinocerotidae* (Rhinoceros), *Papio sp* (Baboon) and *Tragelaphus oryx* (Common eland). *Rhinocerotidae* suffers from very low precision (0.0889) despite a higher recall (0.6667), leading to an F1-score of 0.1569. *Papio sp* similarly displays a low recall of 0.0714, resulting in an F1-score of 0.1333. For *Tragelaphus oryx*, the model achieves a lower precision of 0.8333 and recall of 0.2381, reflecting its inconsistent ability to detect these species accurately. These challenges highlight the model's difficulty with less distinguishable species (caused by night, occlusion or distance) or those underrepresented in the dataset.

**Table 1.** Accuracy, Precision, Recall and F1-Score results for all Sub-Saharan species following Phi3.5-Vision model processing without YOLOv10-X object detection support.

| Class | Common Name | Accuracy | Precision | Recall | F1-Score |
|---|---|---|---|---|---|
| *Canis mesomelas* | Black-backed jackal | 0.9833 | 1.0000 | 0.2000 | 0.3333 |
| *Syncerus caffer* | African buffalo | 0.9833 | 1.0000 | 0.6364 | 0.7778 |
| *Aepyceros melampus* | Impala | 0.9631 | 1.0000 | 0.1818 | 0.3077 |
| *Hippotragus equinus* | Roan antelope | 0.9514 | 1.0000 | 0.0800 | 0.1429 |
| *Bunolagus monticularis* | Bushman rabbit | 0.9553 | 1.0000 | 0.0833 | 0.1538 |
| *Phacochoerus africanus* | Common warthog | 0.9400 | 1.0000 | 0.1667 | 0.2857 |
| *Pan troglodytes* | Chimpanzee | 0.9671 | 1.0000 | 0.2727 | 0.4286 |
| *Smutsia gigantea* | Giant Ground Pangolin | 0.8835 | 1.0000 | 0.0882 | 0.1622 |
| *Orycteropus afer* | Aardvark | 0.9514 | 1.0000 | 0.1429 | 0.2500 |
| *Hippopotamus amphibious* | Common hippopotamus | 0.9553 | 1.0000 | 0.8330 | 0.1538 |
| *Oryx gazella* | South African oryx | 0.9514 | 1.0000 | 0.2000 | 0.3333 |
| *Struthio camelus* | Common ostrich | 0.9833 | 1.0000 | 0.6000 | 0.7500 |
| *Alcelaphus buselaphus* | Hartebeest | 0.9792 | 1.0000 | 0.1667 | 0.2857 |
| *Kobus ellipsiprymnus* | Waterbuck | 0.9833 | 1.0000 | 0.2000 | 0.3333 |
| *Gorilla sp* | Gorilla | 0.9711 | 1.0000 | 0.6818 | 0.8108 |
| *Tragelaphus eurycerus* | Bongo | 0.9216 | 1.0000 | 0.0909 | 0.1667 |
| *Kobus kob* | African Antelope | 0.9792 | 1.0000 | 0.2857 | 0.4444 |
| *Numida meleagris* | Helmeted guineafowl | 0.9874 | 1.0000 | 0.2500 | 0.4000 |
| *Hystrix cristata* | Crested porcupine | 0.9592 | 1.0000 | 0.0909 | 0.1667 |
| *Crocuta crocuta* | Spotted hyena | 0.9833 | 1.0000 | 0.5556 | 0.7143 |
| *Loxodonta africana* | African elephant | 0.9751 | 1.0000 | 0.8636 | 0.9268 |
| *Acinonyx jubatus* | Cheetah | 0.9751 | 1.0000 | 0.6250 | 0.7692 |
| *Papio sp* | Baboon | 0.9004 | 1.0000 | 0.0714 | 0.1333 |
| *Rhinocerotidae* | Rhinoceros | 0.3778 | 0.0889 | 0.6667 | 0.1569 |
| *Connochaetes taurinus* | Blue wildebeest | 0.9363 | 1.0000 | 0.3043 | 0.4667 |
| *Tragelaphus oryx* | Common eland | 0.9325 | 0.8333 | 0.2381 | 0.3704 |
| *Giraffa camelopardalis* | Giraffa | 0.9553 | 1.0000 | 0.7105 | 0.8308 |
| *Panthera leo* | African Lion | 0.7966 | 1.0000 | 0.3103 | 0.4737 |
| *Equus quagga* | Plains zebra | 0.9216 | 1.0000 | 0.5238 | 0.6875 |

The confusion matrix in Figure 11 provides a detailed breakdown of the classifications made by the Phi-3.5 model. As indicated by the confusion matrix and supported by the results in Table 1, the model performed well for certain species. However, other species, show more frequent misclassifications. This analysis provides valuable insights into the strengths and weaknesses of the Phi-3.5 model when detecting and classifying different animal species.

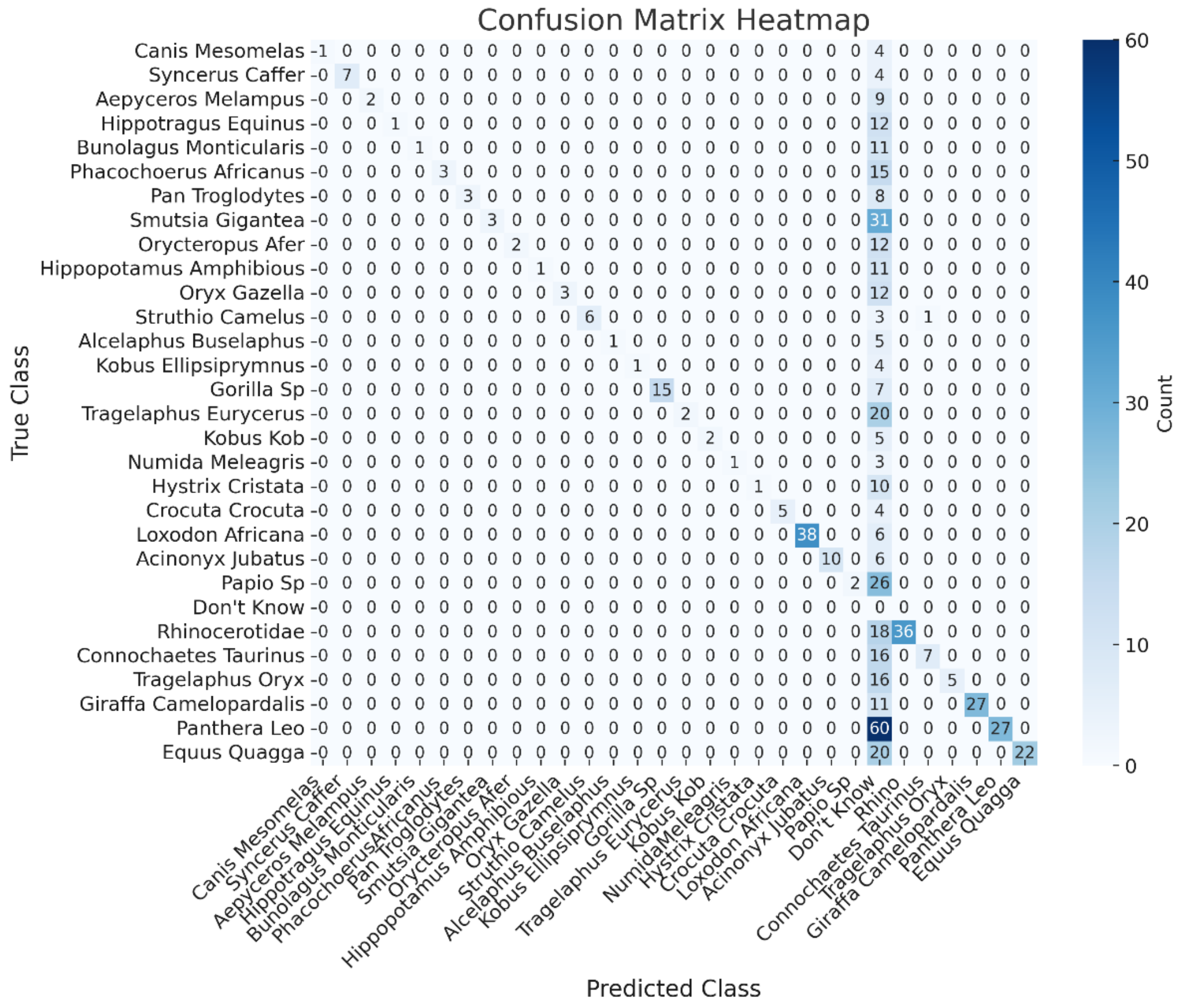

**Figure 10.** The confusion matrix provides a detailed breakdown of the classifications made by the Phi 3.5-vision model when applied to raw images without YOLOv10-X object detection support.

### *4.3. Results for Vision-Language Model with OD Support*

This section presents the performance of the Phi 3.5 model with object detection support for animal identification. The labelled images, from YOLOv10-X, are processed by the Phi-3.5 model, which uses its optical character recognition (OCR) capabilities to identify the animals based on the bounding box text. This two-step method overcomes Phi-3.5'smodel's limitations reported in Section 4.2.

Compared to these results, object detection support significantly reduces misclassifications and increases overall accuracy (Table 2), particularly in cases where the model previously struggled with lower recall and species identification challenges. For example, species like *Tragelaphus eurycerus* (Bongo) and *Papio sp.* (Baboons), which exhibited lower recall and F1-scores, now show improved identification with more balanced precision and recall pairs. Species such as *Hippopotamus amphibius* (Common hippopotamus), *Oryx gazella* (South African oryx), *Alcelaphus buselaphus* (Hartebeest), *Gorilla sp.* (Gorilla), *Kobus kob* (African Antelope), and *Numida meleagris* (Helmeted guineafowl) achieve perfect scores (1.0000) for accuracy, precision, recall, and F1-score, showcasing the model's precise identification capabilities without misclassification. Other species, including *Syncerus caffer* (African buffalo) and *Struthio camelus* (Common ostrich), also perform well, with recall values of 0.8462 and 0.8000, respectively, and F1-scores of 0.9167 and 0.8889. However, certain species, such as *Tragelaphus eurycerus* (Bongo) and *Papio sp.* (Baboon), present challenges, even with object detection support (probably due to animals caught in the camera trap, at distance, or at night). For instance, *Tragelaphus eurycerus* (Bongo) shows a recall of 0.0909 and an F1-score of 0.1667, reflecting ongoing difficulties in reliably identifying this species. Similarly, *Papio sp.* (Baboon) achieves a recall of 0.4444 and an F1-score of 0.6154, indicating challenges in classification for that class. Additionally, for *Rhinocerotidae* (Rhinoceros), the model shows a high recall of 0.9825 but struggles with precision (0.4628), resulting in a modest F1-score of 0.6292. This suggests that while the model captures a large number of true positives for *Rhinocerotidae* (Rhinoceros), it remains prone to misclassifications (likely caused by images captured at night which are more difficult to classify).

Overall, the model with object detection support exhibits excellent performance across most species, particularly those with distinct morphometric characteristics, with many species achieving near-perfect metrics as shown in Table 2.

**Table 2.** Accuracy, Precision, Recall and F1-Score results for all Sub-Saharan species following Phi3.5-Vision model processing with object detection support.

| Class | Common Name | Accuracy | Precision | Recall | F1-Score |
|---|---|---|---|---|---|
| *Canis mesomelas* | Black-backed jackal | 0.9942 | 0.7000 | 1.0000 | 0.8235 |
| *Syncerus caffer* | African buffalo | 0.9961 | 1.0000 | 0.8462 | 0.9167 |
| *Aepyceros melampus* | Impala | 0.9923 | 1.0000 | 0.6364 | 0.7778 |
| *Hippotragus equinus* | Roan antelope | 0.9904 | 1.0000 | 0.5444 | 0.7059 |
| *Bunolagus monticularis* | Bushman rabbit | 0.9981 | 1.0000 | 0.9167 | 0.9565 |
| *Phacochoerus africanus* | Common warthog | 0.9904 | 1.0000 | 0.7222 | 0.8387 |
| *Pan troglodytes* | Chimpanzee | 0.9981 | 0.9167 | 1.0000 | 0.9565 |
| *Smutsia gigantea* | Giant Ground Pangolin | 0.9847 | 1.0000 | 0.6800 | 0.8095 |
| *Orycteropus afer* | Aardvark | 0.9866 | 1.0000 | 0.6111 | 0.7586 |
| *Hippopotamus amphibious* | Common hippopotamus | 1.0000 | 1.0000 | 1.0000 | 1.0000 |
| *Oryx gazella* | South African oryx | 1.0000 | 1.0000 | 1.0000 | 1.0000 |
| *Struthio camelus* | Common ostrich | 0.9961 | 1.0000 | 0.8000 | 0.8889 |
| *Alcelaphus buselaphus* | Hartebeest | 1.0000 | 1.0000 | 1.0000 | 1.0000 |
| *Kobus ellipsiprymnus* | Waterbuck | 0.9942 | 1.0000 | 0.4000 | 0.5714 |
| *Gorilla sp* | Gorilla | 1.0000 | 1.0000 | 1.0000 | 1.0000 |
| *Tragelaphus eurycerus* | Bongo | 0.9626 | 1.0000 | 0.0909 | 0.1667 |
| *Kobus kob* | African Antelope | 1.0000 | 1.0000 | 1.0000 | 1.0000 |
| *Numida meleagris* | Helmeted guineafowl. | 1.0000 | 1.0000 | 1.0000 | 1.0000 |
| *Hystrix cristata* | Crested porcupine | 0.9942 | 1.0000 | 0.7273 | 0.8421 |
| *Crocuta crocuta* | Spotted hyena | 0.9942 | 0.7692 | 1.0000 | 0.8696 |
| *Loxodonta africana* | African elephant | 0.9828 | 0.8431 | 0.9773 | 0.9053 |
| *Acinonyx jubatus* | Cheetah | 0.9923 | 1.0000 | 0.7895 | 0.8824 |
| *Papio sp* | Baboons | 0.9717 | 1.0000 | 0.4444 | 0.6154 |
| *Rhinocerotidae* | Rhinoceros | 0.8864 | 0.4628 | 0.9825 | 0.6292 |
| *Connochaetes taurinus* | Blue wildebeest | 0.9904 | 0.9091 | 0.8696 | 0.8889 |
| *Tragelaphus oryx* | Common eland | 0.9923 | 1.0000 | 0.8095 | 0.8947 |
| *Giraffa camelopardalis* | Giraffa | 0.9961 | 0.9737 | 0.9737 | 0.9737 |
| *Panthera leo* | African Lion | 0.9828 | 0.9333 | 0.9655 | 0.9492 |
| *Equus quagga* | Plains zebra | 0.9981 | 0.9756 | 1.0000 | 0.9877 |

The confusion matrix (Figure 12) offers a comprehensive breakdown of the Phi-3.5 model's performance with OD support. The matrix highlights that the model performs exceptionally well for several species, with most true instances aligning along the diagonal - indicative of correct classifications. For example, *Canis mesomelas* (Black-backed jackal) has 7 true positives, and *Syncerus caffer* (African buffalo) has 11 true positives, showing minimal confusion with other species.

Despite these successes, some misclassifications are observed, particularly with species such as *Papio sp.* (Baboon), *Phacochoerus africanus* (Common warthog), and *Smutsia gigantea* (Giant Ground Pangolin). For instance, *Papio sp.* (Baboon) is occasionally misclassified as *Acinonyx jubatus* (Cheetah) or given a "Don't Know" response. Similarly, *Phacochoerus africanus* (Common warthog) is sometimes incorrectly classified as *Syncerus caffer* (African buffalo). Additionally, *Smutsia gigantea* (Giant Ground Pangolin) exhibits slight confusion with *Canis mesomelas* (Black-backed jackal) in certain cases.

Overall, the confusion matrix illustrates that while the model achieves high accuracy for many species, certain species remain challenging to classify. The integration of object detection has reduced some of these errors, but there is still room for improvement, particularly in handling species that are more difficult to distinguish.

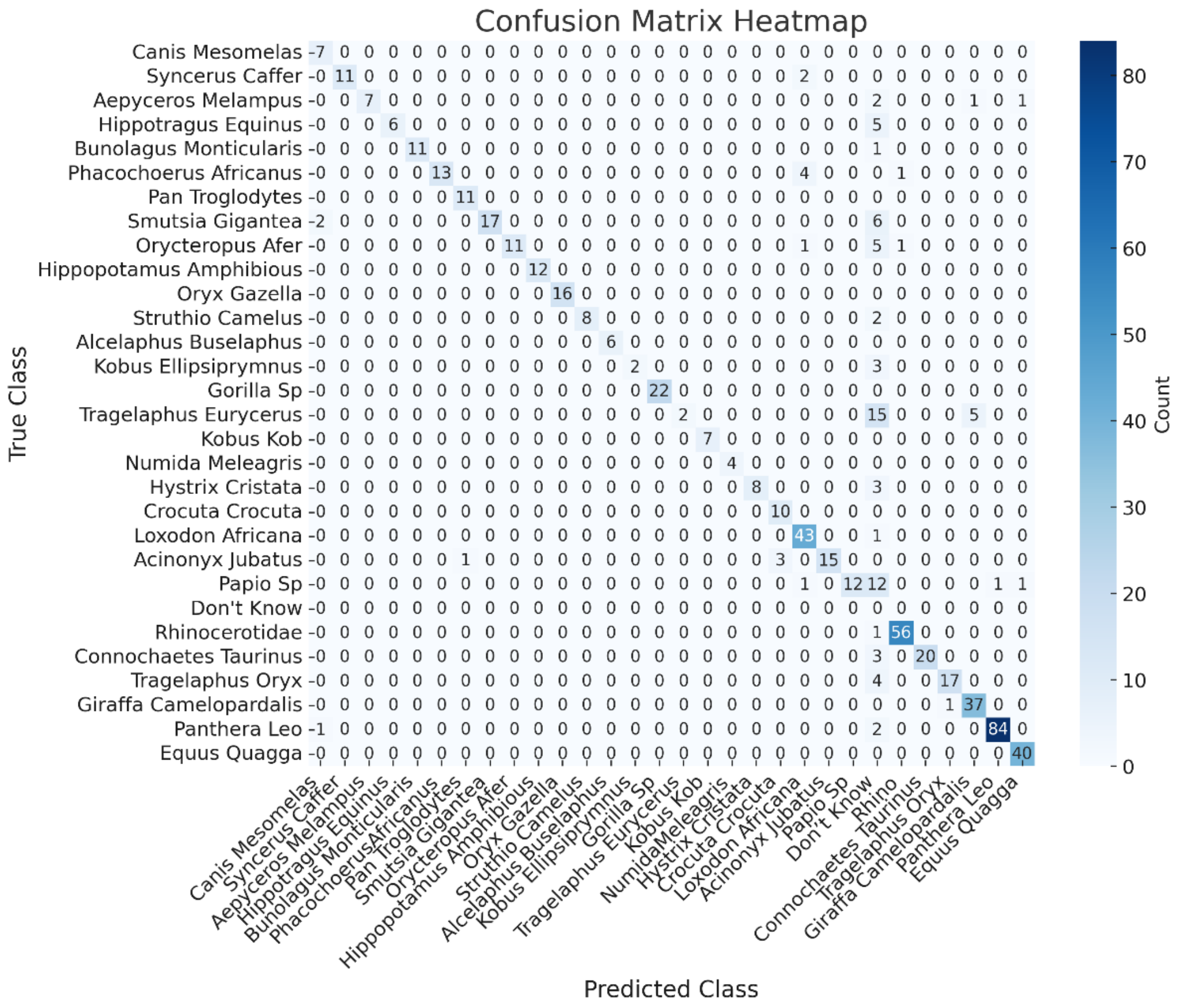

**Figure 11.** Confusion matrix for the Phi 3.5 model using the od binding boxes from the test case images.

### *4.4. Results for Retrieval Augmented Generation*

We now evaluate the capabilities of the Phi-3.5 model within the RAG framework. Building on its success in species identification with object detection support, we now assess how the model synthesises detailed information about the identified species using external knowledge and compares its outputs with ground truth responses (provided by conservationists).

The results presented (see Table 3 and Appendix A) show the BERTScores (precision, recall, and F1-score) and the images used to evaluate the similarity between the model-generated answers and the ground truth. These scores assess the contextual relevance and accuracy of the model's answers, as compared to the ground truth. For instance, the question "*Was the image taken during the day or night?*" achieved high precision (0.9469), recall (0.9165), and F1-score (0.9165), indicating that the model was highly effective in interpreting time and environmental factors. Similarly, for questions related to species identification and IUCN conservation status, the model produced strong results, with an F1-score of 0.9382, underscoring its ability to accurately retrieve and present relevant species-specific information from external sources.

More complex reasoning or comparison-based questions, such as "How does the species identified in the image compare to other species in the same habitat?" (F1-score: 0.8685) and "What are the known predators or threats to the species?" (F1-score: 0.8585), also demonstrated strong performance. By focusing on the comparison between the generated answers and ground truth, the evaluation highlights the system's ability to deliver accurate and reliable information across a wide range of ecological and environmental questions. The variation in scores - particularly between fact-based questions and those requiring broader ecosystem-level insights (e.g., "What is the species' role in the ecosystem?" F1-score: 0.8554)—shows that while the system performs well, additional refinement may be needed for more complex, multi-faceted inquiries.

**Table 3.** Phi-3.5 and RAG Q&A Results.

| Question | Answer | Ground Truth | Precision | Recall | F1-Score |
|---|---|---|---|---|---|
| **Q1. Read the label on the binding box to identify the animal. What is the species identified in the image, and what is its IUCN conservation status?** | A1. The species identified in the image is a rhino. According to the provided passages, the IUCN conservation status of the rhinoceros is "Near Threatened. | GT1. The species in the image is a rhinoceros and its IUCN status is near threatened. | 0.9572 | 0.9199 | 0.9382 |
| **Q2 Read the label on the binding box to identify the animal. What is the average weight of the species identified, and does this species have any notable characteristics or behaviours?** | A2. The animal identified in the image is a *Connochaetes taurinus*, also known as the blue wildebeest. The average weight of the blue wildebeest is around 170 to 410 kg (370 to 900 lb). This species is known for its distinctive blue-grey skin, which is unique among wildebeest species. They are also known for their migratory behaviour, forming large herds that travel in search of food and water. | GT2. There are two Connochaetes taurinus in the image. The average weight is 250 kg (550 lb). It is highly gregarious and superbly adapted for a migratory existence. | 0.8798 | 0.8530 | 0.8662 |
| **Q3. Was the image taken during the day or night, and what environmental factors can be observed (e.g., forest, bush, water sources)?** | A3. The image was taken during the day. The environmental factors that can be observed include a grassy hill with some bushes and trees scattered around. There are no visible water | GT3. The image was taken during the day. The environment is a grassland with several trees typical of an African plain. | 0.9469 | 0.9165 | 0.9165 |
| | sources in the image. | | | | |
| **Q4. Read the label on the binding box to identify the animal. How does the species identified in the image compare to other species in the same habitat in terms of size, behaviour, and diet?** | A4. The species identified in the image is Giraffa camelopardalis, which is a large herbivorous mammal native to Africa. Giraffes are known for their long necks and legs, and they are the tallest mammals on Earth. They are typically found in savannas, grasslands, and open woodlands in Africa. Giraffes are also known for their unique behavior of browsing on leaves, flowers, and fruits from tall trees, which sets them apart from other antelope species that primarily graze on grasses. In terms of diet, giraffes are herbivores and primarily feed on leaves, flowers, and fruits from trees, particularly from the Acacia tree. | GT4. The animal is a Giraffa camelopardalis. Giraffa camelopardalis are the tallest mammal with a long neck that makes it ideal for eating the leaves and twigs of acacia, mimosa, and wild apricot trees. Giraffa camerlopardalis are herbivores. | 0.8738 | 0.8632 | 0.8685 |
| **Q5. Read the label on the binding box to identify animals. Can you identify other animals or objects in the image, such as nearby trees, water bodies, or structures?** | A5. In the image, there is a zebra standing in the center of the frame. There are no other animals visible in the image. The background consists of a dirt road and some vegetation, but no water bodies or | GT5. The animal in the image is a Equus quagga. It is standing on a dirt track which is surrounded by trees. There are no other animals in the image. | 0.8992 | 0.8966 | 0.8979 |
| | structures are visible. | | | | |
| **Q6 Read the labels on the binding boxes to identify animals. What animals are in the image and how many are there of each animal species identified?** | A6. The image contains three rhinoceroses. | GT6. The animals in the image are rhinoceroses. There are three of them. | 0.8764 | 0.8783 | 0.8774 |
| **Q7. Based on the species and its habits, what predictions can be made about its activity at the time the camera trap image was taken (e.g., hunting, foraging, resting)?** | A7. Based on the species and its habits, it can be predicted that the lion is likely resting or inactive at the time the camera trap image was taken. Lions spend much of their time resting, with only intermittent bursts of activity throughout the day. The image shows a lion lying down on the grass, which supports this prediction. | GT7. The picture contains several lions – only one is fully visible. All the lions are lying down resting. | 0.8977 | 0.8581 | 0.8774 |
| **Q8. Read the label on the binding box around the animal to determine the species. What potential threats, either natural or human-induced, are most relevant to the species in the image, given its current IUCN status and environment?** | A8. The species in the image is a pangolin, specifically the Giant pangolin (Manis gigantea). Given its current IUCN status as endangered, the potential threats most relevant to the species in the image include illegal, international trade in pangolin skin, scales, and meat, as well as habitat loss and fragmentation. These threats are | GT8. The image contains a pangolin which is the most hunted mammal with an IUCN status of endangered. Its main threat is humans who poach them for the meat and scales. | 0.9088 | 0.8712 | 0.8896 |
| | primarily human-induced and can lead to a decline in pangolin populations. | | | | |
| **Q9. Read the label on the binding box around the animal to determine the species. What is the species role in the ecosystem, and how does its presence effect other species or the environment in the area where the image was captured?** | A9. The species in the image is the African bush elephant (Loxodonta africana). As a keystone species, the presence of African bush elephants in the ecosystem has a significant impact on other species and the environment. They play a crucial role in shaping the landscape by uprooting trees and undergrowth, which can transform savannah into grasslands. This creates more accessible habitat for smaller herbivores to access trees that have been mowed down by elephants. Additionally, their digging for water during dry seasons can create water holes that benefit other species in the area. | GT9. The image contains an African elephant. They maintain habitats and promote the growth of larger trees. They trample forests and dense grasslands, making room for smaller species to co-exist. | 0.8727 | 0.8388 | 0.8554 |
| **Q10. Read the label on the binding box around the animal to determine the species. What are the known predators or threats to the** | A10. The species in the image is a spotted hyena (Crocuta crocuta). Known predators or threats to the spotted hyena include lions (Panthera leo), cheetahs (Acinonyx | GT10. The animal in the image is Crocuta Crocuta. Its only true predators are lions and humans. There are no visible threats in the image. | 0.9034 | 0.8178 | 0.8585 |

| | |
|---|---|
| **species in the image, and are there any visible indicators in the environment that suggest the presence of these threats?** | jubatus), leopards (Panthera pardus), brown hyenas (Parahyaena brunnea), and African wild dogs (Lycaon). There are no visible indicators in the environment that suggest the presence of these threats. |

*4.5. Automated Reporting*

To streamline the analysis of camera trap data, an automated reporting system was developed. This system integrates the outputs of YOLOv10-X and Phi-3.5 models to efficiently generate structured reports. After images are processed by the Phi-3.5 model, the Llama-2-7b-hf model is employed to automatically generate a set of questions and corresponding answer pairs using the Phi-3.5 derived information. This process leverages the Alpaca format for question-answer pair generation, ensuring consistency and clarity in the presentation of the extracted data (see Figure 12).

```
{
  "instruction": "What are the environmental factors observed in the image with the blue wildebeest?",
  "input": "",
  "output": "The environmental factors observed in the image include a grassy hill with some bushes and trees
  scattered around. There are no visible water sources in the image.",
  "metadata": {
    "heading": "Environmental Factors: Blue Wildebeest Image",
    "date": "2024-10-23"
  }
},
```

**Figure 12.** Alpaca JSON format showing the question-answer pairs.

This function is applied to the entire dataset of camera trap images, resulting in a comprehensive collection of species-specific analyses. The use of the Alpaca format enables the integration of this information into downstream analytical workflows or direct reporting (see Figure 13 for sample report).

In order to further enhance the utility of these outputs, the Alpaca-formatted data is converted into a structured report using the Python docx library. This conversion process translates the JSON data into a well-organised Word document that is accessible to interested stakeholders. The report includes information on species identification, conservation status, environmental factors, and behavioural predictions, from all camera traps, providing a holistic view of each observation. Note that this is a simplistic example that demonstrates the applicability of the approach and not a useful report suitable for stakeholder tasks.

# Wildlife Species Report

Generated on: 2024-10-23

## Introduction

This report contains detailed information about various wildlife species identified through images using the Phi-3.5-vision-instruct model. The following sections outline species details, their conservation status, and observed environmental factors.

## Species Identification and Conservation Status: Rhinoceros

The species identified in the image is a rhinoceros. According to the provided passages, the IUCN conservation status of the rhinoceros is 'Near Threatened.'

Date: 2024-10-23

## Species Information: Blue Wildebeest

The species identified in the image is Connochaetes taurinus, also known as the blue wildebeest. The average weight of the blue wildebeest is around 170 to 410 kg (370 to 900 lb). This species is known for its distinctive blue-grey skin, which is unique among wildebeest species.

Date: 2024-10-23

## Environmental Factors: Blue Wildebeest Image

The environmental factors observed in the image include a grassy hill with some bushes and trees scattered around. There are no visible water sources in the image.

Date: 2024-10-23

**Figure 12.** Sample report using Alpaca Q&A.

## 5. Discussion

The results from this study highlight both the strengths and weaknesses of the Phi-3.5 model, particularly in challenging scenarios involving low-quality camera trap images. In the initial set of experiments, where the model processed images without object detection support, it became evident that identifying species in such conditions is inherently difficult (likely as the model was not initially trained on camera trap images). While the model demonstrated high precision for certain specie - such as *Syncerus caffer* (African Buffalo) and *Struthio camelus* (Common ostrich), which achieved perfect precision (1.0000; Table 1) - it struggled significantly with recall. For instance, *Canis mesomelas* (Black backed jackal) had a recall of just 0.2000, leading to a low F1-score of 0.3333 (Table 1), underscoring the model's difficulty in consistently identifying species without localisation assistance.

With the integration of object detection, the performance of the Phi-3.5 model improved significantly across all key metrics. By using the YOLOv10-X model to localise and classify animals first within images and combining the results with the optical character recognition capabilities of the Phi-3.5 model for species identification, the system achieved substantial gains. For example, the F1-score for *Canis mesomelas* rose sharply from 0.3333 to 0.8235 (Table 2), once including our multi-modal approach. Similarly, *Syncerus caffer*, which already performed well, saw its F1-score improve to 0.9167. Notably, some species, such as *Hippopotamus amphibius* and *Oryx gazella*, achieved perfect scores (1.0000) across accuracy, precision, recall, and F1-score, underscoring the effectiveness of combining object detection with vision-language models.

By focusing on interpreting labels within bounding boxes, the Phi-3.5 model bypassed many of the challenges associated with direct image analysis, significantly reducing the rate of misclassification and improving accuracy. Additionally, the model could extract peripheral information, such as environmental features like trees and water sources, as well as metadata such as time stamps and camera IDs.

Another less obvious benefit of incorporating the YOLO model is its low inferencing cost and its ability to remove blank images. Since blank images make up approximately 68% of camera trap datasets, removing these using the YOLOv10-X model allows for a more efficient and cost-effective solution, reducing computational overheads and enabling faster processing of meaningful data. Note: the 68% of blanks a the time of writing is based on the number of images processed by Conservation AI and the number of observations we have seen (35,018,212 images processed and 11,024,671 animals detected in those images).

However, one major issue encountered during the study was the inconsistent presentation of bounding boxes and text labels. Non-standard colour combinations, such as white text on pink backgrounds, made it difficult for the OCR component to accurately read the labels, leading to species misidentifications. Additionally, certain images - especially higher resolution ones - featured thinner bounding boxes and smaller text, which further complicated label readability. In these instances, the Phi 3.5 model struggled with text recognition, resulting in misclassifications and missed identifications. For example, even with object detection, *Rhinocerotidae* continued to present challenges, achieving a precision of 0.4628 and an F1-score of 0.6292 (Table 2), likely due to these text readability issues.

Despite the improvements from object detection, certain species - such as *Papio sp.* and *Tragelaphus eurycerus* - continued to exhibit relatively low performance (the later images were of a much higher resolution). While *Papio sp.* showed some improvement, it only achieved a recall of 0.4444 and an F1-score of 0.6154, reflecting ongoing challenges in accurate species identification. Similarly, *Tragelaphus eurycerus* had a recall of 0.0909 and an F1-score of 0.1667, underscoring the model's persistent difficulty in reliably classifying these species, particularly those with fewer samples or more ambiguous visual characteristics.

Building on the successful integration of the two models, the use of RAG further demonstrated the systems capacity to further enrich species identification by incorporating external contextual information. By sourcing data from Wikipedia, the model provided supplementary insights such as average weight, conservation status, and environmental context. This extended the model's functionality beyond simple species identification, adding significant ecological and biological depth to the analysis. Using the Q&A feature over this integration F1-scores ranged between 0.82 and 0.94 for various answer-ground-truth comparisons, for example, its precision for identifying species and their IUCN conservation status reached 0.9572, with a recall of 0.9199 and an F1-score of 0.9382. Similarly, its handling of environmental factors, like determining whether an image was taken during the day or night, yielded an F1-score of 0.9165. These scores indicate that the RAG-enhanced model can accurately combine object detection with species identification and contextual knowledge, enriching the system's overall output.

From a conservation perspective, this system demonstrates significant potential in streamlining workflows and improving data accessibility for stakeholders. The ability to integrate species identification with contextual information, such as conservation status and environmental context, could inform better decision-making for habitat protection and species management. For instance, the identification of invasive species or degraded habitats from camera trap images could help conservationists allocate resources more efficiently. However, further work is needed to align these outputs with conservation priorities, such as automated biodiversity metrics, population density estimations, and species movement tracking. Building on this proof of concept, the system could be enhanced with tailored datasets and integration into larger conservation workflows, ultimately supporting real-time monitoring and preventative conservation actions.

## 6. Conclusions

This study enhances camera trap image analysis by integrating advanced AI models. The YOLOv10-X object detection model enabled precise animal detection and localisation, while the Phi-3.5-vision-instruct model incorporated vision-language capabilities for species identification and extraction of peripheral environmental data. Additionally, the integration of RAG further enriched the system by retrieving detailed species-specific information, such as IUCN status, average weight, and environmental context, from external knowledge sources like Wikipedia.

This combined approach demonstrates significant improvements in species classification, particularly in challenging low-quality images where traditional models often fail. The high accuracy, precision, recall, and F1-scores across most species validate the effectiveness of the methodology. Moreover, the RAG component added additional contextual richness by providing supplementary insights, which are critical for informed wildlife management decisions. The system's potential integration with emerging technologies, such as drone-based monitoring or satellite imagery, also paves the way for broader conservation applications, enhancing the utility of this framework.

The inclusion of an automated reporting system, while rudimentary in its current implementation, demonstrates the potential for automatically generating structured reports based on model outputs. By providing stakeholders with immediate access to information, this system could significantly reduce manual effort. By continuing to refine the system, this integrated AI approach offers a scalable, efficient, and cost-effective solution to wildlife conservation. It provides deeper insights, enabling more timely and effective conservation efforts on a global scale. Ultimately, the combination of object detection, vision-language models, and RAG offers a transformative advancement in wildlife monitoring and species management.

However, despite the encouraging results, several challenges remain. Inconsistent presentation of bounding boxes and text labels, as well as difficulties in detecting species with ambiguous or low-quality visual features, limited the system's performance in specific cases, such as with *Papio sp.* and *Rhinocerotidae*. Additionally, readability issues due to non-standard text formats and small bounding boxes were identified as areas for improvement. Future work will continue to evaluate this system further and investigate the optimal use of AI model composition to help improve the results reported in this paper.

One promising future direction involves a more streamlined approach where, instead of passing the image with bounding boxes and labels to the Phi-3.5 model for species identification, the system could extract this information from a structured database (e.g., a SQL record). By directly informing the Phi-3.5 model of the presence and identity of the species, the model could then focus solely on providing additional contextual information using its internal vision capabilities. This approach could bypass the limitations of text readability and bounding box inconsistencies. Implementing such a solution would require advanced prompt engineering, an area not covered extensively in this study, which employed only basic prompt techniques. Further investigation into optimised prompt engineering could greatly improve the system's overall performance and facilitate more accurate contextual analysis.

**Appendix A**

In Appendix A, we include a set of 10 images corresponding to each of the questions evaluated in RAG Q&A Table 3. These images provide visual context for the species and environmental factors discussed in the questions, allowing for a clearer understanding of the challenges and outcomes presented. Each image represents a real-world example of the scenarios encountered in a study, from species identification and environmental observations to the retrieval-augmented generation of supplementary information like conservation status and ecosystem roles. By including these visuals, we aim to further support the quantitative results with qualitative examples that illustrate the effectiveness and limitations of the proposed system in processing and analysing camera trap images.

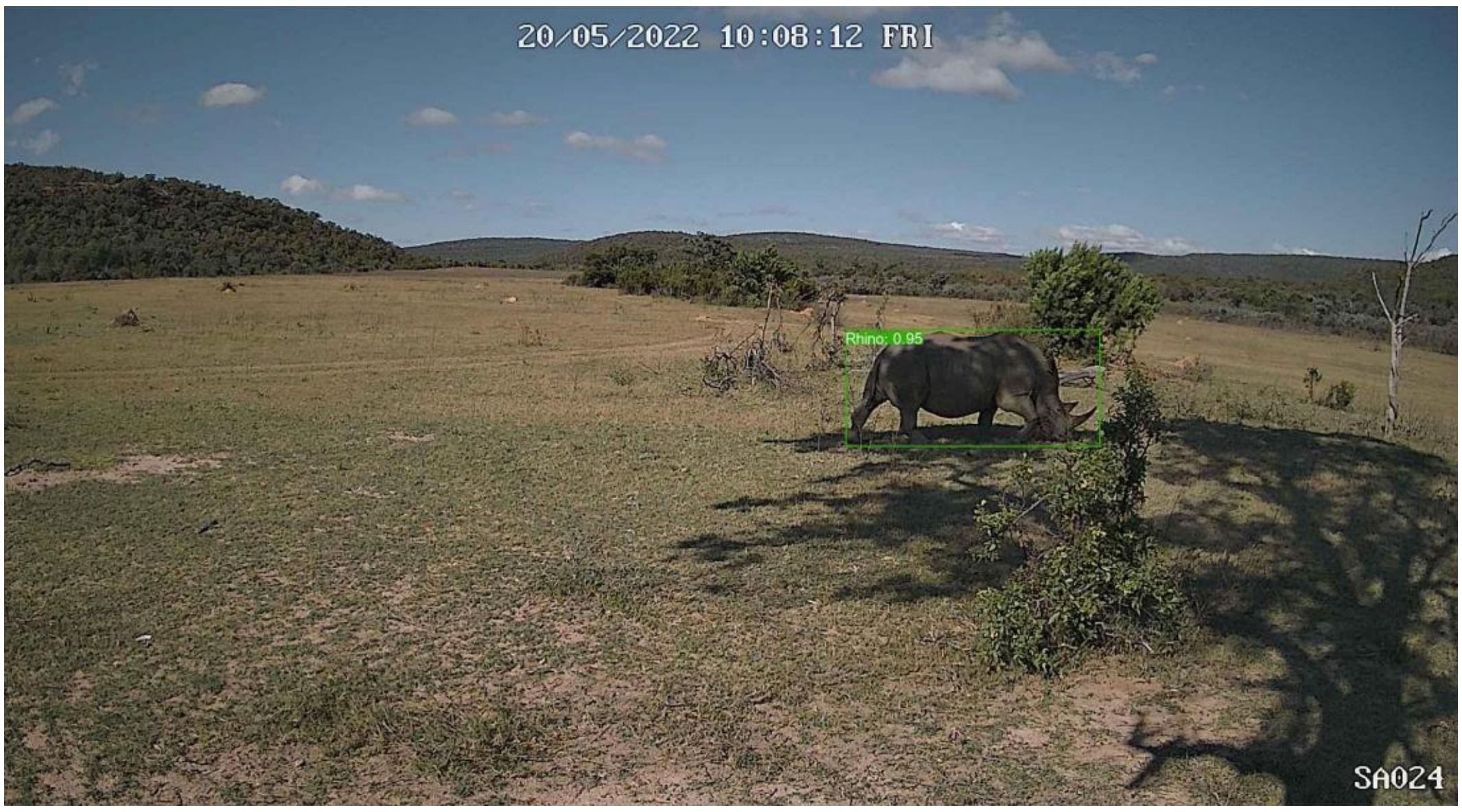

**Figure A1.** Q1. Read the label on the binding box to identify the animal. What is the species identified in the image, and what is its IUCN conservation status?

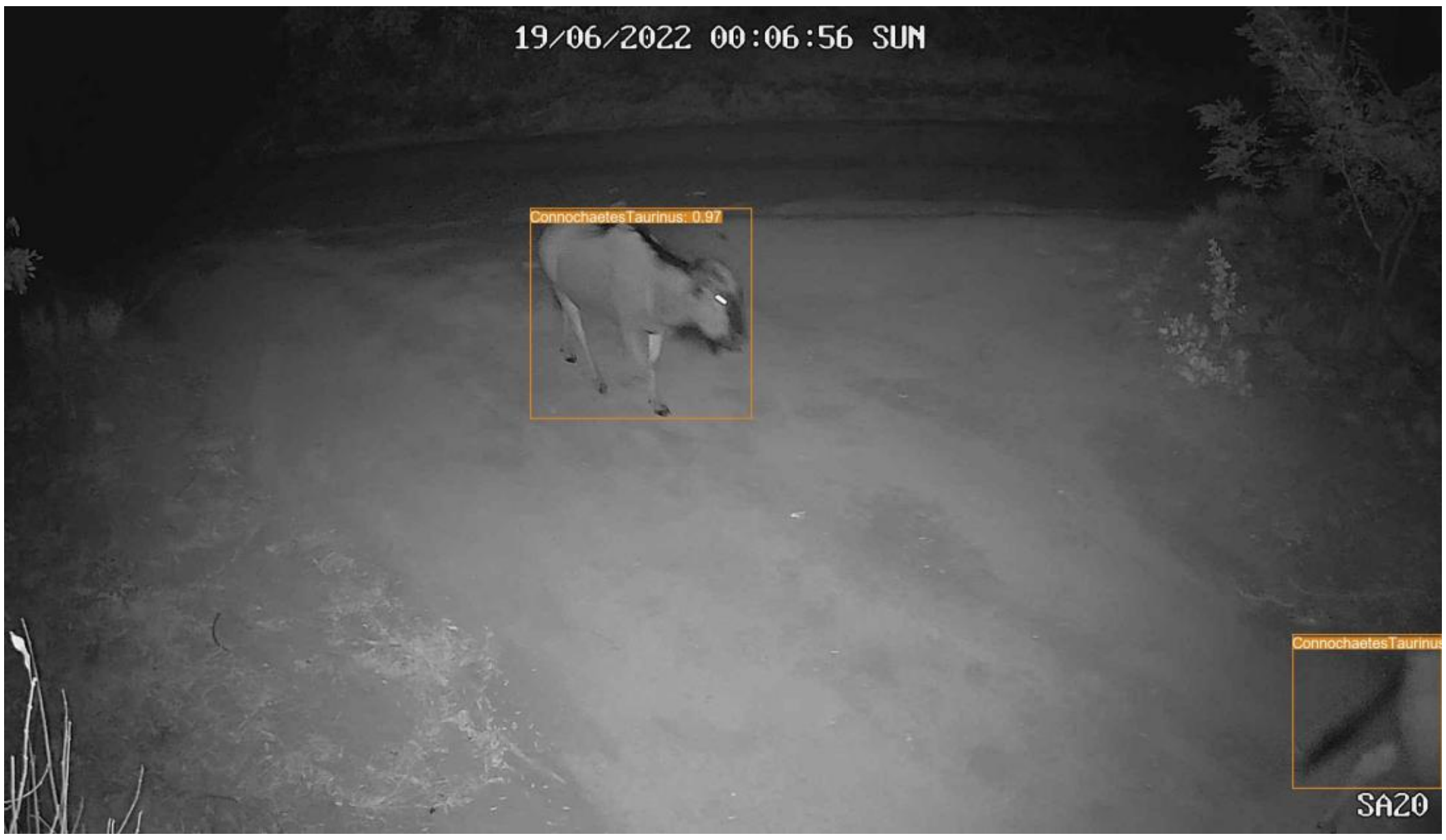

**Figure A2.** Q2 Read the label on the binding box to identify the animal. What is the average weight of the species identified, and does this species have any notable characteristics or behaviours?

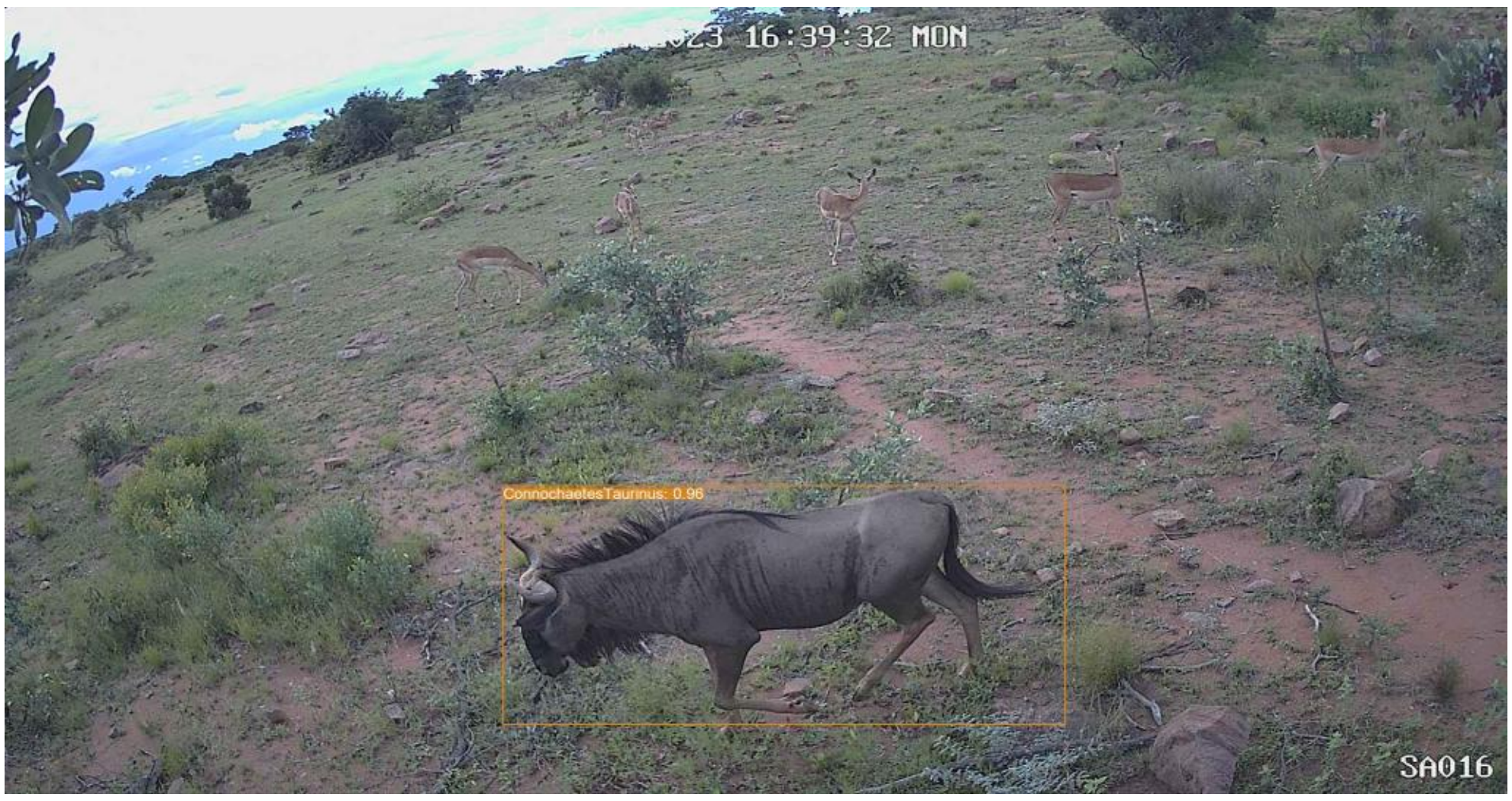

**Figure A3. Q3.** Was the image taken during the day or night, and what environmental factors can be observed (e.g., forest, bush, water sources)?

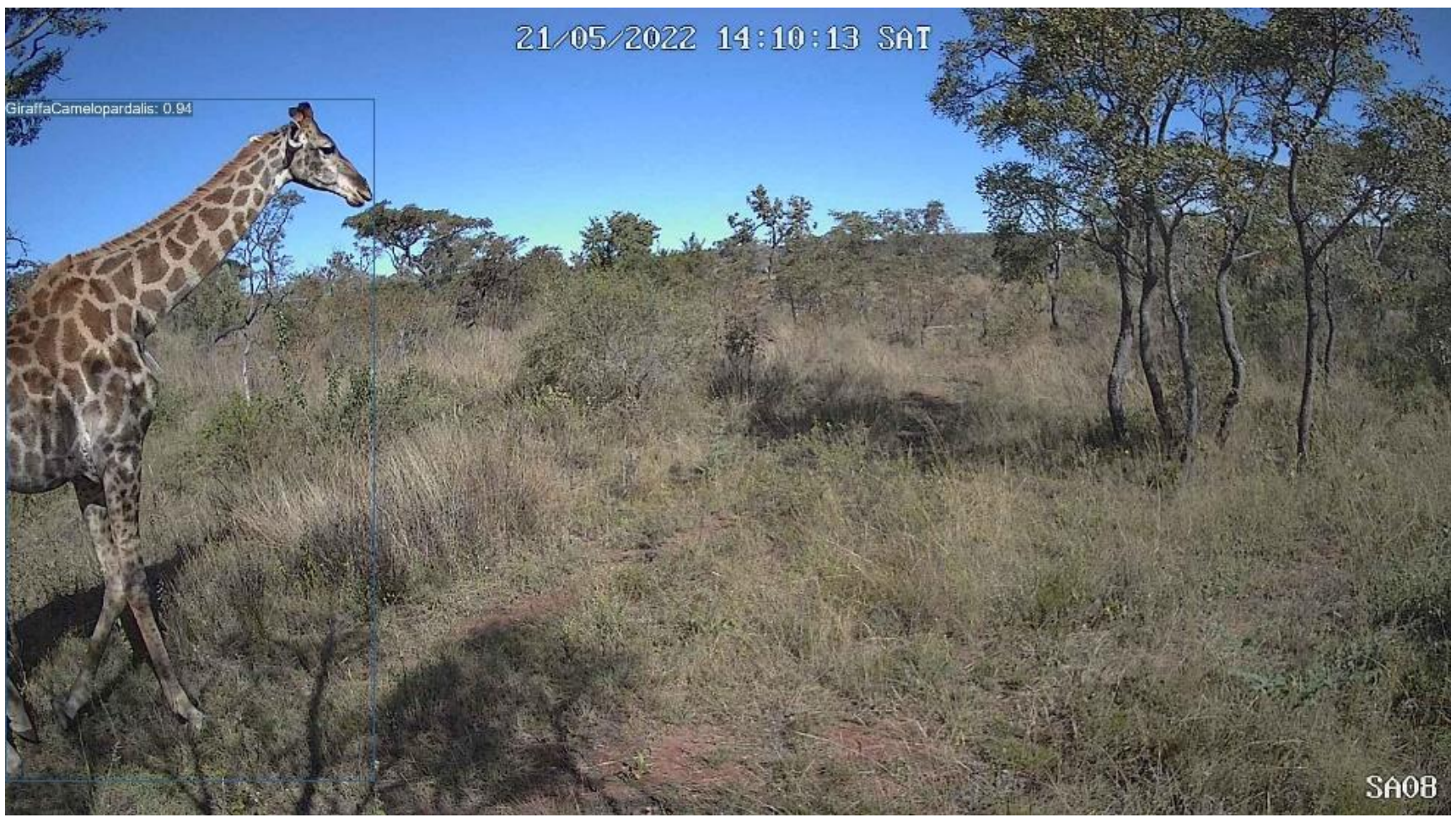

**Figure A4. Q4.** Read the label on the binding box to identify the animal. How does the species identified in the image compare to other species in the same habitat in terms of size, behaviour, and diet?

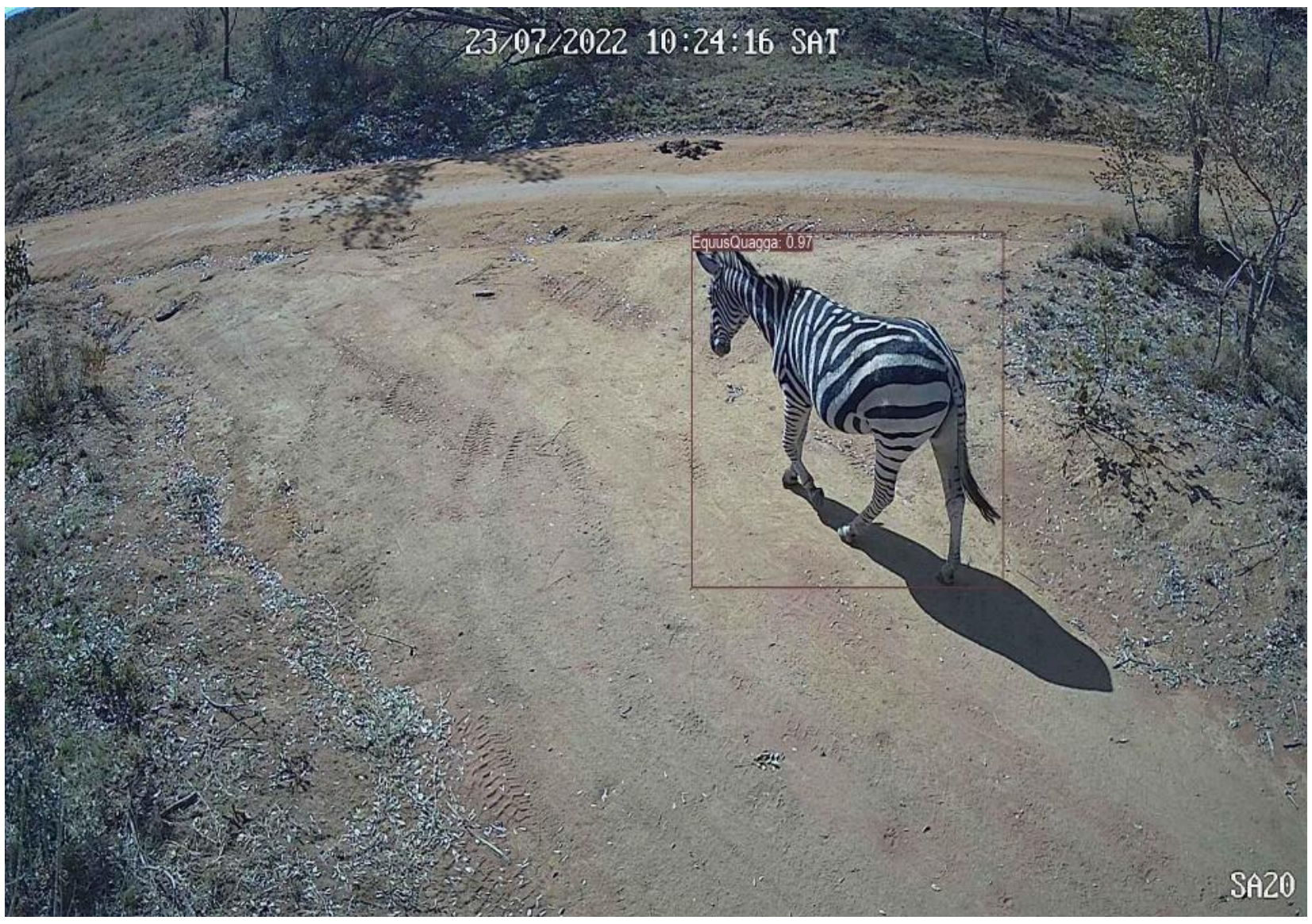

**Figure A5.** Q5. Read the label on the binding box to identify animals. Can you identify other animals or objects in the image, such as nearby trees, water bodies, or structures?

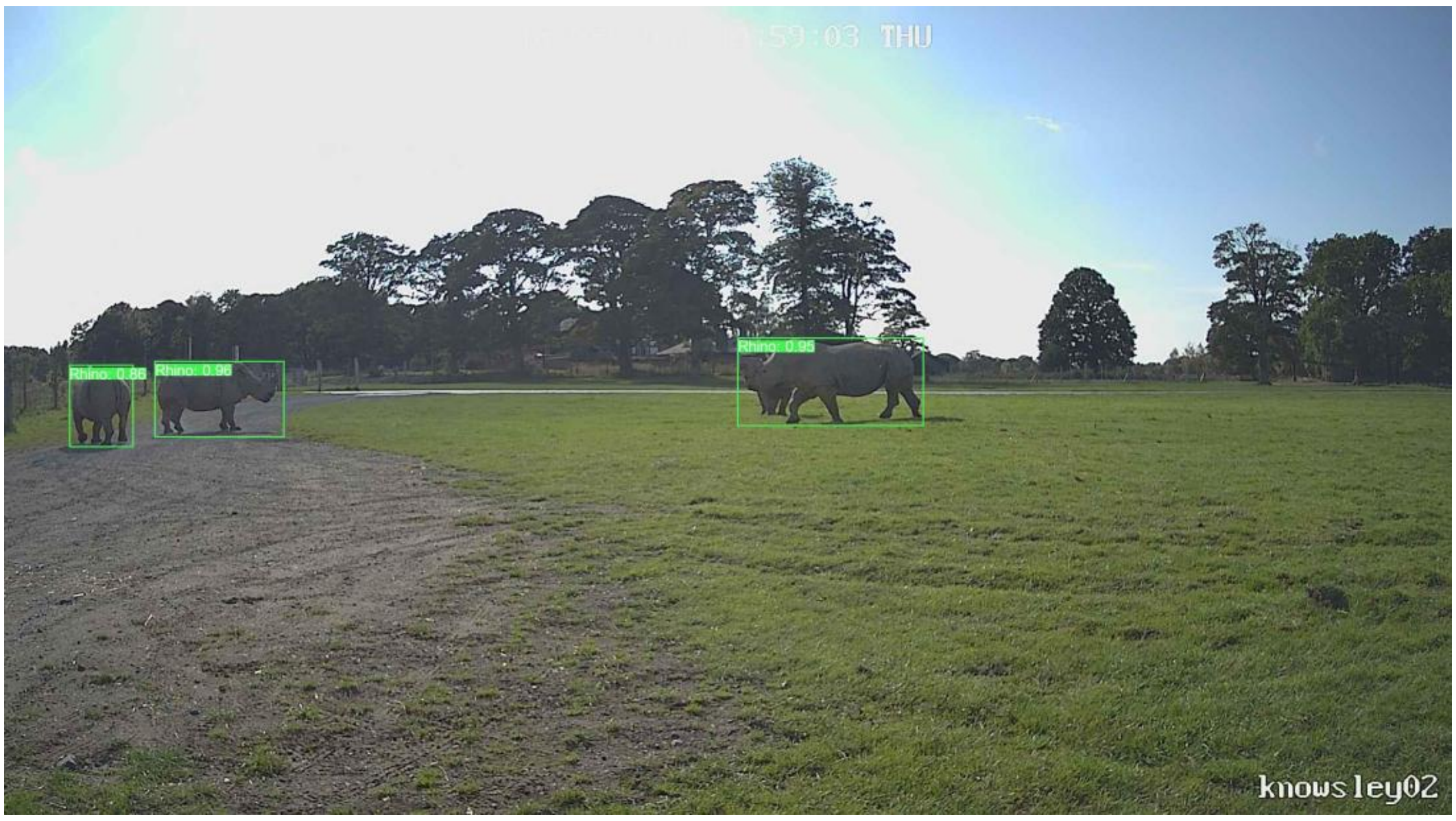

**Figure A6.** Q6 Read the labels on the binding boxes to identify animals. What animals are in the image and how many are there of each animal species identified?

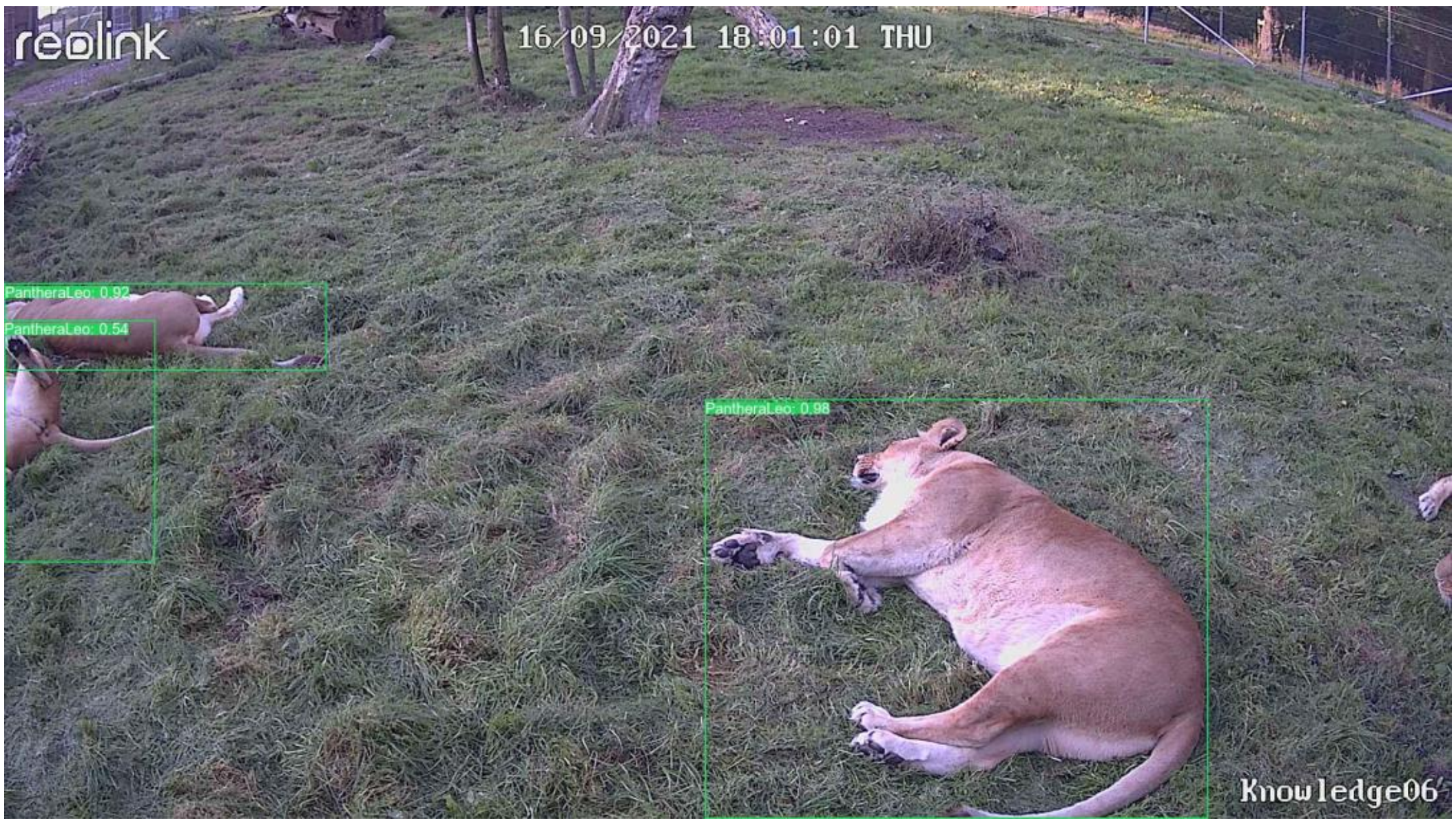

**Figure A7.** Q7. Based on the species and its habits, what pre-dictions can be made about its activity at the time the camera trap image was taken (e.g., hunting, foraging, resting)?

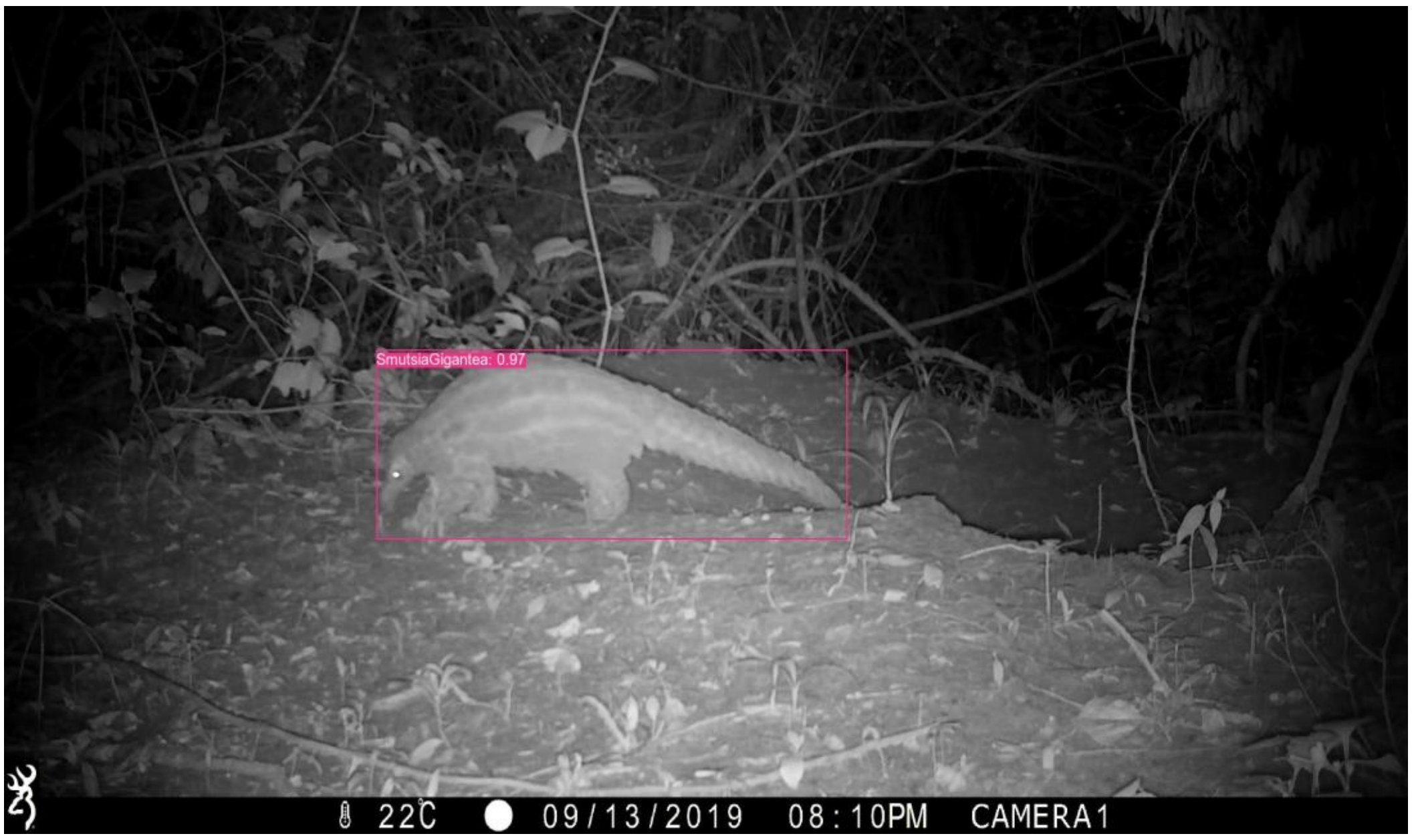

**Figure A8.** Q8. Read the label on the binding box around the animal to determine the species. What potential threats, either natural or human-induced, are most relevant to the species in the image, given its cur-rent IUCN status and environment?

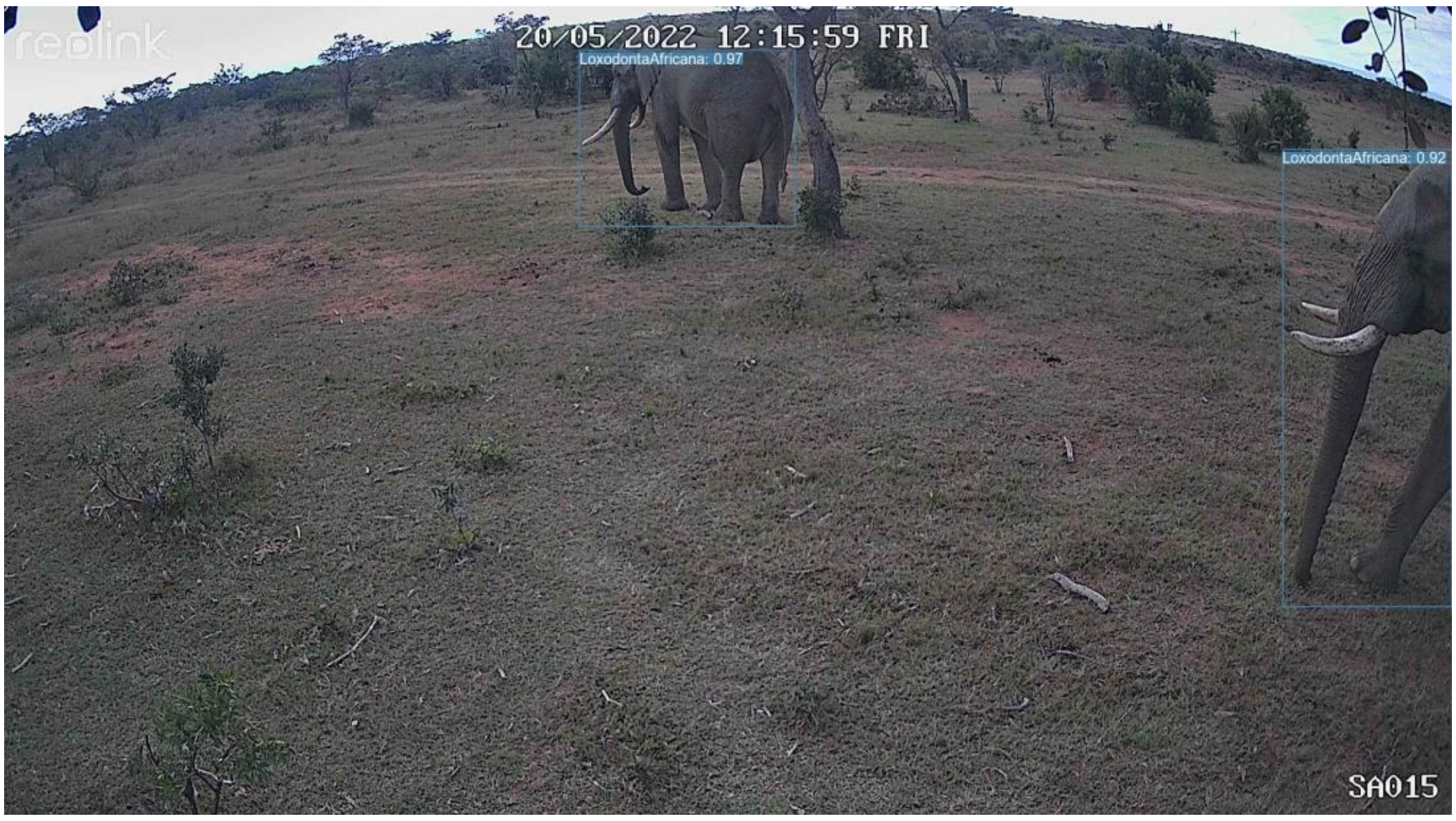

**Figure A9.** Q9. Read the label on the binding box around the animal to determine the species. What is the species role in the ecosystem, and how does its presence effect other species or the environment in the area where the image was captured?

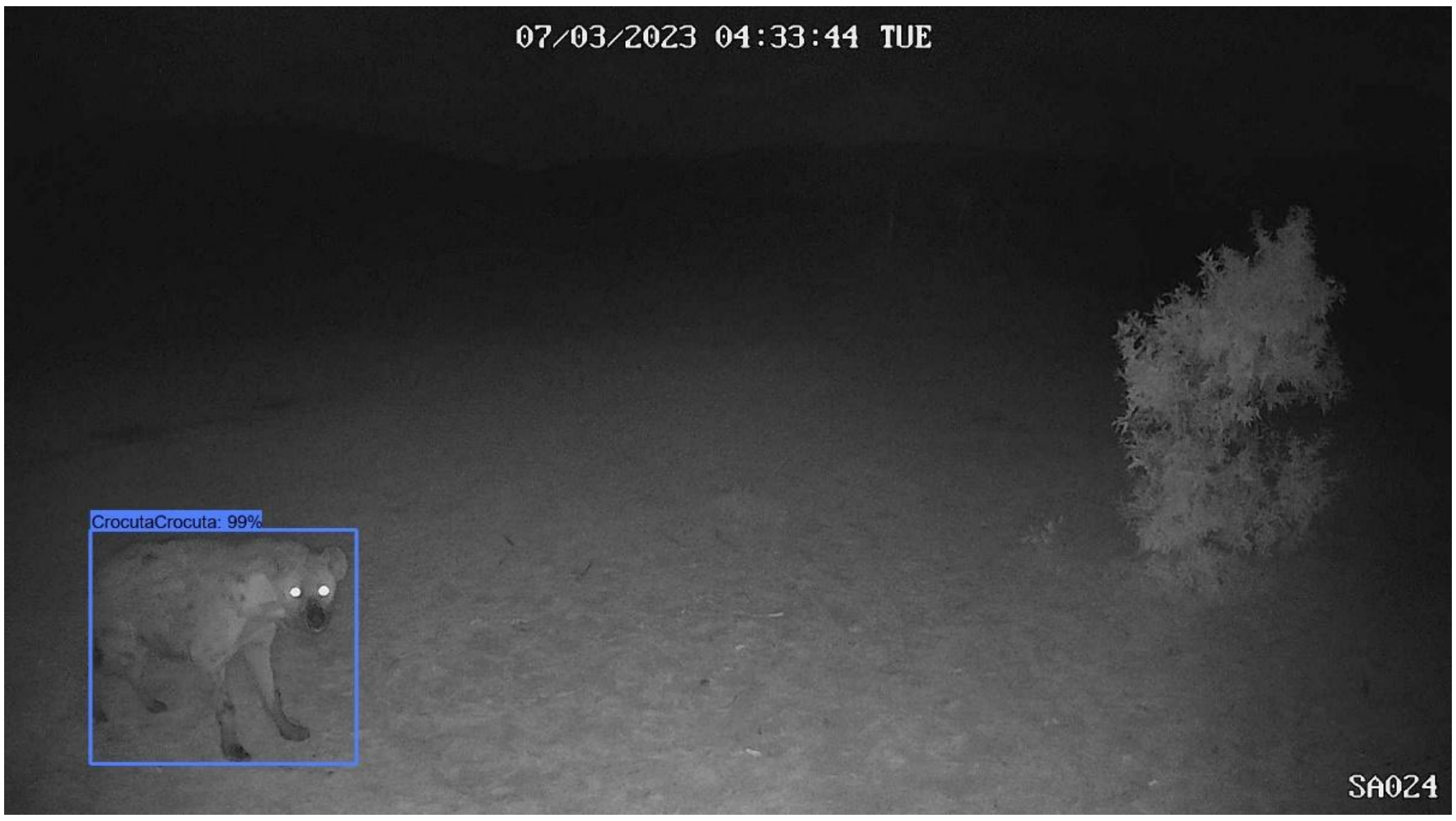

**Figure A10.** Q10. Read the label on the binding box around the animal to determine the species. What are the known predators or threats to the species in the image, and are there any visible indicators in the environment that suggest the presence of these threats?